\documentclass[10pt,twocolumn,letterpaper]{article}
\usepackage[pagenumbers]{cvpr} % To force page numbers, e.g. for an arXiv version

\usepackage[accsupp]{axessibility}
\usepackage[dvipsnames]{xcolor}
\usepackage{amsmath,amssymb,amsfonts}
\usepackage{algorithmic}
\usepackage{graphicx}
\usepackage{textcomp}
\usepackage{xcolor}
\usepackage{color}
\usepackage{colortbl}
\usepackage{mathtools}
\usepackage{amsmath}
\usepackage{tabularx}
\usepackage{multirow}
\usepackage{multicol}
\usepackage{float}
\usepackage{tabularx}
\usepackage{threeparttable}
\usepackage{booktabs}
\usepackage{longtable}
\usepackage{makecell}
\usepackage{verbatim}
\usepackage{wrapfig}
\usepackage{xspace}
\usepackage{pbox}
\usepackage{epstopdf}
\usepackage{comment}
\usepackage{lipsum}
\usepackage{bm}
\usepackage{bbm}
\usepackage{setspace}
\usepackage{sidecap}
\usepackage{array}
\usepackage{blindtext}
\usepackage{enumitem}
\usepackage{diagbox}
\usepackage{bbding}
\usepackage{url}
\usepackage{indentfirst}
\definecolor{cvprblue}{rgb}{0.21,0.49,0.74}
\usepackage[pagebackref,breaklinks,colorlinks,citecolor=cvprblue]{hyperref}

\usepackage{svg}

\usepackage[capitalize]{cleveref}
\crefname{section}{Sec.}{Secs.}
\crefname{section}{Section}{Sections}
\crefname{table}{Table}{Tables}
\crefname{table}{Tab.}{Tabs.}

\newcommand{\benchmark}{Mimicking-Bench\xspace}

\def\paperID{498} % *** Enter the Paper ID here
\def\confName{CVPR}
\def\confYear{2025}

\title{Mimicking-Bench: A Benchmark for Generalizable Humanoid-Scene \\ Interaction Learning via Human Mimicking}

\author{
Yun Liu\textsuperscript{*,1,2,3,4},~
Bowen Yang\textsuperscript{*,1,2},~
Licheng Zhong\textsuperscript{*,3},~
He Wang\textsuperscript{2,5},~
Li Yi\textsuperscript{\textdagger,1,3,4}
\smallskip\\
\textsuperscript{1}Tsinghua University~~~
\textsuperscript{2}Galbot~~~
\textsuperscript{3}Shanghai Qi Zhi Institute\\
\textsuperscript{4}Shanghai Artificial Intelligence Laboratory~~~
\textsuperscript{5}Peking University \\
\url{https://mimicking-bench.github.io/}
}

\begin{document}

\twocolumn[{
\renewcommand\twocolumn[1][]{#1}
\maketitle
\begin{center}
    \captionsetup{type=figure}
    \includegraphics[width=1.0\textwidth]{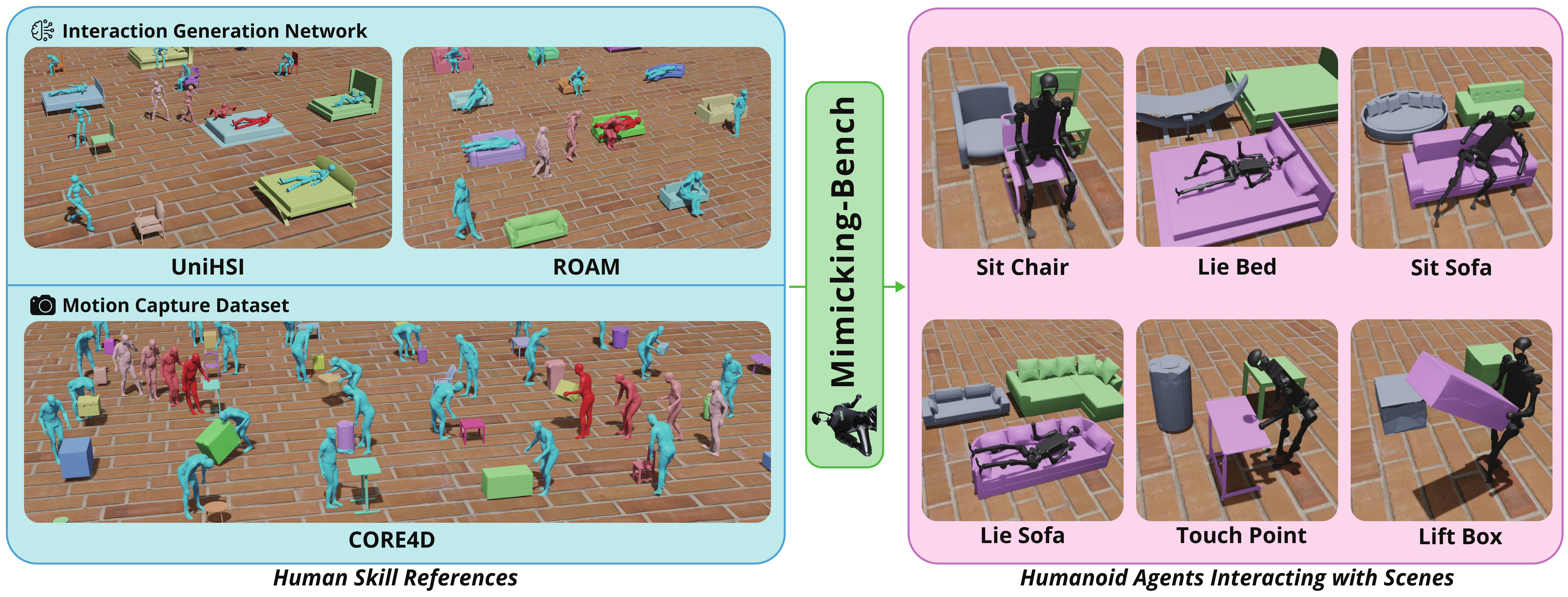}
    \caption{\textbf{\benchmark} is the first benchmark for learning generalizable humanoid-scene interaction skills via mimicking human data, comprising six household interaction tasks. It integrates a diverse human skill reference dataset by leveraging the advances in motion capture dataset and interaction generation network, and constructs a skill-learning paradigm for human-to-humanoid knowledge transfer.}
    \label{fig:teaser}
\end{center}
}]

\footnotetext[1]{Equal contribution.}
\footnotetext[2]{Corresponding author.}

\begin{abstract}
Learning generic skills for humanoid robots interacting with 3D scenes by mimicking human data is a key research challenge with significant implications for robotics and real-world applications. However, existing methodologies and benchmarks are constrained by the use of small-scale, manually collected demonstrations, lacking the general dataset and benchmark support necessary to explore scene geometry generalization effectively. To address this gap, we introduce \benchmark, the first comprehensive benchmark designed for generalizable humanoid-scene interaction learning through mimicking large-scale human animation references. \benchmark includes six household full-body humanoid-scene interaction tasks, covering 11K diverse object shapes, along with 20K synthetic and 3K real-world human interaction skill references. We construct a complete humanoid skill learning pipeline and benchmark approaches for motion retargeting, motion tracking, imitation learning, and their various combinations. Extensive experiments highlight the value of human mimicking for skill learning, revealing key challenges and research directions.

\end{abstract}

\section{Introduction}

The past decade has seen remarkable advancements in humanoid robotics~\cite{unitree_h1,feng2014optimization,boston_atlas} and the development of skill learning algorithms for physical simulation environments~\cite{tang2024humanmimic,zhang2024achieving} as well as real-world deployments~\cite{meng2023online,TRILL,liao2024berkeley,chen2024learning,zhuang2024humanoid,gu2024humanoid}. For instance, humanoids can now generalize walking skills across a variety of previously unseen terrains by leveraging diverse training data~\cite{radosavovic2024humanoid}. Meanwhile, there have been significant strides in robot manipulation skills~\cite{zhang2024wococo,merel2020catch,dao2024sim}, and the emerging task of learning generalizable humanoid skills for interacting with complex scenes is gaining traction. This is a critical step forward for deploying humanoids in real-world scenarios, where they can assist in tasks traditionally requiring human intervention, thereby reducing labor costs and enhancing efficiency.

% Due to complicated humanoid dynamics, leveraging reinforcement learning that explores environments from scratch faces significant challenges in learning generalizable interaction skills that can adapt to diverse scene and object geometries. Benefitting from topological similarities between humanoid and human skeletons, a promising alternative is regarding versatile human skill data as a rich knowledge source and inducing humanoids to mimic them~\cite{he2024omnih2o,fu2024humanplus}. Despite being intuitive, this learning paradigm involves several technical difficulties to overcome, including eliminating differences between humanoid and human models, transferring skill animations to executable humanoid control signals, and emerging general policies from numerous skill references.

Due to the complex dynamics of humanoid robots, applying reinforcement learning~\cite{ppo,hansen2023td} to explore environments from scratch presents significant challenges in learning generalizable interaction skills that can adapt to a wide variety of scene and object geometries. A promising alternative leverages the topological similarities between humanoid and human skeletons, using versatile human skill data as a rich source of knowledge and guiding humanoids to mimic these skills~\cite{he2024omnih2o,fu2024humanplus}. While intuitive, this approach faces several technical hurdles, including addressing the discrepancies between humanoid and human models, translating skill animations into executable humanoid control signals, and deriving generalized policies from a diverse set of skill references.
% Previous technical advances~\cite{he2024omnih2o,fu2024humanplus,ze2024generalizable} present solutions to parts of these challenges, however, they are confined to several specific interaction scenarios due to costly real-world robot hardware usage. Therefore, the research community always demands benchmarks~\cite{mu2021maniskill,zhu2020robosuite,james2020rlbench,yu2020meta} to establish platforms to compare various algorithms systematically and promote synergistic studies. However, existing benchmarks for humanoid-scene interaction either exclude utilizing human knowledge~\cite{sferrazza2024humanoidbench} or limit to small-scale specific human demonstrations collected by expensive teleoperation~\cite{chernyadev2024bigym}, which can hardly support comprehensive studies on mimicking diverse human skill references.
% ~\eric{the logic is not smooth here} However, the research community lacks comprehensive benchmarks to promote synergistic studies on these challenges, ~\eric{the logic is not smooth here either} where 
Previous works~\cite{he2024omnih2o,fu2024humanplus,ze2024generalizable} have tackled parts of these challenges, but they are limited to specific interaction scenarios due to the high costs of real-world robot hardware. As a result, there is a growing demand for comprehensive benchmarks~\cite{mu2021maniskill,zhu2020robosuite,james2020rlbench,yu2020meta} that provide standardized platforms for comparing various algorithms and promoting collaborative research. However, existing benchmarks for humanoid-scene interactions either exclude the use of human knowledge~\cite{sferrazza2024humanoidbench} or focus on small-scale, specific demonstrations collected through expensive teleoperation~\cite{chernyadev2024bigym}, making it hard to conduct systematic studies on mimicking a diverse set of human skill references.

% To fill this research gap, we present \benchmark, the first benchmark for generalizable humanoid-scene interaction learning by mimicking large-scale diverse human skill references, encompassing six household tasks in whole-body contact scenarios. Constructing \benchmark could face two main challenges: how to collect extensive human skill references that cover interaction with various object geometries, and which technical aspects in this paradigm deserve further study.
% what are exactly the technical challenges in human mimicking~\eric{the second challenge is not clear here}.
To fill this research gap, we present \benchmark, the first benchmark for generalizable humanoid-scene interaction learning through the mimicking of large-scale, diverse human skill references. \benchmark spans six household tasks involving whole-body contact scenarios. Constructing \benchmark addresses two key challenges: first, how to collect a diverse set of human skill references that cover interactions with a wide range of object geometries; and second, identifying which technical aspects within this paradigm warrant further investigation and development.
% 
% Benefitting from versatile human-scene interaction datasets~\cite{SAMP,CHAIRS,core4d} and derived interaction generation networks~\cite{unihsi,roam,zhang2022couch} in computer vision and graphics fields, we address the first challenge by integrating real-world and automatically synthesized human-scene interactions as a comprehensive skill knowledge source, resulting in 23K human-scene interaction motion sequences spanning 11K diverse object geometries.
% ~\eric{we should emphasize on what we mean by generalizable somewhere before so that the object number does not look weird here}.
Benefiting from versatile human-scene interaction datasets~\cite{SAMP,CHAIRS,core4d} and derived interaction generation networks~\cite{unihsi,roam,zhang2022couch} in the fields of computer vision and graphics, we address the first challenge by integrating both real-world and automatically synthesized human-scene interaction data. This comprehensive skill knowledge source encompasses 23K human-scene interaction motion sequences, covering 11K diverse object geometries.
% 
% To handle the second challenge, we investigate various existing human mimicking pipelines~\cite{fu2024humanplus,he2024omnih2o,tang2024humanmimic,tessler2024maskedmimic} and summarize three successive core technical modules: retargeting human motions to humanoid skeletons~\cite{he2024omnih2o,li2024okami}, tracking skill animations in physical simulation environments~\cite{tessler2024maskedmimic,he2024hover}, and imitating humanoid skill demonstrations~\cite{act,diffusion_policy,liu2024opt2skill}. By connecting these modules, \benchmark constructs a general skill-learning paradigm covering different existing pipelines. With vast algorithms for each module, \benchmark supports two types of studies: 1) comparing existing prevailing humanoid skill learning pipelines~\cite{he2024omnih2o,fu2024humanplus}, and 2) evaluating different outstanding modular algorithms to promote comprehensive studies on all key techniques.
% 
To address the second challenge, we investigate various existing human mimicking pipelines~\cite{fu2024humanplus,he2024omnih2o,tang2024humanmimic,tessler2024maskedmimic} and identify three successive core technical modules: 1) retargeting human motions to humanoid skeletons~\cite{he2024omnih2o,li2024okami}, 2) tracking skill animations in physical simulation environments~\cite{tessler2024maskedmimic,he2024hover}, and 3) imitating humanoid skill demonstrations~\cite{act,diffusion_policy,liu2024opt2skill}. By connecting these modules, \benchmark constructs a general skill-learning paradigm that encompasses different existing pipelines. With a wide array of algorithms available for each module, \benchmark enables two types of studies: 1) comparing existing, well-established humanoid skill learning pipelines~\cite{he2024omnih2o,fu2024humanplus}, and 2) evaluating various outstanding modular algorithms, promoting comprehensive research on all key techniques involved in humanoid-scene interaction learning.

% Extensive experiments are conducted in pipeline-wise and module-wise method comparisons, indicating the importance of proper modular algorithm selection. Integrating the best algorithm of each module, mimicking human references acquires more natural motions and significantly enhances task success rates compared to knowledge-free reinforcement learning, demonstrating its effectiveness and potential.
Extensive experiments are conducted through pipeline-wise and module-wise method comparisons, highlighting the critical importance of selecting the appropriate algorithms for each module. By integrating the best algorithm for each module, the approach of mimicking human references results in more natural motion generation and significantly improves task success rates compared to data-free reinforcement learning. These results demonstrate the effectiveness of human mimicking in humanoid-scene interaction learning and its potential to advance the field.
% \modify{summarize from experiment results} ~\eric{any other interesting findings? data scaling? more?}

% In summary, our main contributions are fourfold: 1) We present \benchmark, the first comprehensive benchmark for generalizable humanoid-scene interaction learning via human mimicking. 2) We integrate a large-scale diverse human skill reference dataset comprising synthesized and real-world human-scene interactions. 3) We construct a general skill-learning paradigm and support pipeline-wise and modular evaluations. 4) Experiments demonstrate the effectiveness of the human mimicking paradigm and reveal new challenges and research opportunities.
In summary, our main contributions are fourfold: 1) We present \benchmark, the first comprehensive benchmark for generalizable humanoid-scene interaction learning via human mimicking. 2) We integrate a large-scale and diverse human skill reference dataset that encompasses both synthesized and real-world human-scene interactions. 3) We develop a general skill-learning paradigm and provide support for both pipeline-wise and modular evaluations. 4) Extensive experiments validate the effectiveness of the human mimicking paradigm and reveal new challenges and research opportunities for future exploration.

\section{Related Work}
\subsection{Benchmarks for Humanoid Skill Learning}

Early works~\cite{memmesheimer2019simitate,yu2020meta,james2020rlbench,zhu2020robosuite,heo2023furniturebench,jia2024towards,mu2021maniskill,gu2023maniskill2} have benchmarked robot-object interaction skill learning on dexterous hands~\cite{adroit_hand} and robot arms~\cite{franka}. Driven by rapid progress in humanoid robot models~\cite{kojima2015development,feng2014optimization,unitree_h1} and skills~\cite{radosavovic2024humanoid,ze2024generalizable,zhang2024wococo} in both simulation and real-world scenarios, various benchmarks for humanoid robots have been developed to explore the capabilities of diverse skill learning methods in a scalable and reproducible manner. Table~\ref{tab:benchmarks} provides a brief summary of these benchmarks. For humanoid skill learning, one branch of benchmarks~\cite{brockman2016openai,tassa2018deepmind,al2023locomujoco} focuses on humanoid locomotion skills such as walking, running, and stabilization. Another emerging branch aims to foster studies on humanoid-scene interactions~\cite{sferrazza2024humanoidbench,chernyadev2024bigym}, where HumanoidBench~\cite{sferrazza2024humanoidbench} supports demonstration-free reinforcement learning (RL), and BiGym~\cite{chernyadev2024bigym} collects a small set of demonstrations via manual teleoperation for benchmarking imitation learning methods.
However, these benchmarks are limited by the restricted generalizability of RL algorithms and the small scale of demonstrations due to the high cost of teleoperation. They cannot assess a method's ability to generalize across scene and object geometry changes, nor can they support studies on human-to-humanoid skill transfer. In contrast, \benchmark presents a comprehensive solution by enabling learning from large-scale, diverse human skill references, benchmarking popular algorithms for each core technical module, and supporting research on generalization with diverse object geometries.

\begin{table*}[h!]
\centering
\scriptsize
\addtolength{\tabcolsep}{-3pt}
{
\begin{tabular}{|c|cccc|ccc|ccc|}
\hline
\multirow{2}{*}{Benchmark} & \multicolumn{4}{c|}{Characteristics:} & \multicolumn{3}{c|}{Benchmark Aspects:} & \multicolumn{3}{c|}{\# Number of:} \\
& Bipedal & Interaction & Generalization & Human Reference & Retargeting & Tracking & Imitation Learning & Human Reference & Object & Task \\
\hline
\textit{Plappert et.al.} \cite{plappert2018multi} & & \checkmark & & & & & & - & 3 & 4 \\
MetaWorld \cite{yu2020meta} & & \checkmark & & & & & & - & 66 & 50 \\
RLBench \cite{james2020rlbench} & & \checkmark & & & & & \checkmark & - & 302 & \textbf{106} \\
RoboSuite \cite{zhu2020robosuite} & & \checkmark & & & & & & - & 23 & 9 \\
FurnitureBench \cite{heo2023furniturebench} & & \checkmark & & & & & \checkmark & - & 8 & 8 \\
D3IL \cite{jia2024towards} & & \checkmark & & & & & \checkmark & - & 19 & 7 \\
ManiSkill \cite{mu2021maniskill} & & \checkmark & \checkmark & & & & \checkmark & - & 162 & 4 \\
ManiSkill2 \cite{gu2023maniskill2} & & \checkmark & \checkmark & & & & \checkmark & - & 2.0K & 20 \\
Gym-Mujoco \cite{brockman2016openai} & \checkmark & & & & & & & - & 0 & 2 \\
\textit{Tassa et.al.} \cite{tassa2018deepmind} & \checkmark & & & & & & & - & 0 & 6 \\
LocoMujoco \cite{al2023locomujoco} & \checkmark & & & \checkmark & & & \checkmark & 0.9K & 3 & 27 \\
HumanoidBench \cite{sferrazza2024humanoidbench} & \checkmark & \checkmark & & & & & & - & 28 & 27 \\
BiGym \cite{chernyadev2024bigym} & \checkmark & \checkmark & & & & & \checkmark & - & 71 & 40 \\
\hline
\textbf{\benchmark} & \checkmark & \checkmark & \checkmark & \checkmark & \checkmark & \checkmark & \checkmark & \textbf{23.4K} & \textbf{10.6K} & 6 \\
\hline
\end{tabular}
}
\vspace{-0.2cm}
\caption{\textbf{Comparison of \benchmark with existing robotic skill learning benchmarks.}}
\vspace{-0.4cm}
\label{tab:benchmarks}
\end{table*}

% \subsection{Human to Humanoid, pipeline, retargeting - tracking - IL}

\subsection{Human-Scene Interaction Dataset and Motion Generation}

% Capturing~\cite{su2022robustfusion,xie2022chore,xie2023visibility,xie2024intertrack} and synthesizing~\cite{NIFTY,CHOIS} human-scene and human-object interactions are popular tasks in computer vision and graphics communities. Various dataset efforts collect real-world human motions interacting with static 3D scenes~\cite{SAMP,CHAIRS,PROX,EgoBody}, dynamic objects~\cite{BEHAVE,InterCap,NeuralDome,FORCE,zhao2024m}, and human collaborators~\cite{HOH,core4d,InteractiveHumanoid}, while a recent work~\cite{CIRCLE} presents collecting synthetic interaction data via VR devices. Supported by these datasets, numerous interaction motion generation approaches have emerged with different focuses. A branch of works~\cite{roam,NSM,NIFTY,COUCH,TRUMANS} regards fixed static scenes as inputs and generates human motion only, while another branch~\cite{InterDiff,OMOMO,CHOIS,HOI-Diff,CG-HOI} synthesizes motions of humans and movable objects simultaneously. With rapid progress in large language model (LLM)~\cite{achiam2023gpt}, recent works~\cite{unihsi,InterDreamer,HUMANISE} unlock long-range interaction generation ability by designing LLM planners. Leveraging these advances, we construct a large-scale versatile dataset of human-scene interaction skill references and build up the bridge between motion generation techniques and humanoid skill learning.

Capturing~\cite{su2022robustfusion,xie2022chore,xie2023visibility,xie2024intertrack} and synthesizing~\cite{NIFTY,CHOIS} human-scene and human-object interactions are key tasks in the computer vision and graphics communities. Various dataset initiatives have been launched to collect real-world human motions interacting with static 3D scenes~\cite{SAMP,CHAIRS,PROX,EgoBody}, dynamic objects~\cite{BEHAVE,InterCap,NeuralDome,FORCE,zhao2024m}, and human collaborators~\cite{HOH,core4d,InteractiveHumanoid}, while a recent work~\cite{CIRCLE} explores the collection of synthetic interaction data using VR devices. Supported by these datasets, a variety of motion generation techniques have emerged, each with a different focus. One branch of works~\cite{roam,NSM,NIFTY,COUCH,TRUMANS} treats fixed static scenes as inputs and focuses on generating human motions only, while another branch~\cite{InterDiff,OMOMO,CHOIS,HOI-Diff,CG-HOI} generates motions for both humans and movable objects simultaneously.
With the rapid progress in large language models (LLMs)~\cite{achiam2023gpt}, recent works~\cite{unihsi,InterDreamer,HUMANISE} have unlocked long-range interaction generation abilities by designing LLM planners. By leveraging these advancements, we construct a large-scale, versatile dataset of human-scene interaction skill references and establish a bridge between motion generation techniques and humanoid skill learning.

\subsection{Humanoid Imitation Learning}
\label{sec:related_work_IL}

As a key technical module for humanoids mimicking humans, humanoid imitation learning has been a long-standing research topic aimed at enhancing humanoid skill learning through demonstrations. Considerable progress has been made in various areas, including humanoid locomotion~\cite{tessler2023calm,peng2022ase,tang2024humanmimic,dou2023c,cui2024anyskill,pulse,radosavovic2024humanoid,rempe2023trace,wang2024pacer+}, humanoid-object interaction~\cite{fu2024humanplus,merel2020catch,tessler2024maskedmimic,wang2024skillmimic,unihsi,pan2024synthesizing}, and multi-agent collaboration~\cite{gao2024coohoi,luo2024smplolympics}. In the context of direct solutions, single-stage approaches~\cite{radosavovic2024humanoid,chernyadev2024bigym,tessler2024maskedmimic} use a single policy network to generate actions, where motion discriminators~\cite{gan}, which encode prior knowledge from demonstrations, are incorporated in some works~\cite{peng2021amp,hassan2023synthesizing,tang2024humanmimic,cui2024anyskill,gao2024coohoi,luo2024smplolympics,unihsi,pan2024synthesizing,rempe2023trace,wang2024pacer+} to further improve the naturalness of humanoid motions.

To enable the reuse of learned skills for unseen tasks~\cite{peng2022ase}, two-stage methods~\cite{fu2024humanplus,wang2024skillmimic,tessler2023calm} decompose task fulfillment into two parts: first, planning key motion objectives (such as target joint positions~\cite{fu2024humanplus}, motion semantics~\cite{wang2024skillmimic}, or learnable motion features~\cite{tessler2023calm,peng2022ase,dou2023c,merel2020catch,pulse,juravsky2022padl}), and then controlling the humanoid to complete them. These methods assign the planning and control tasks to a high-level planner and a low-level controller, respectively.

\section{\benchmark}
% We contribute \benchmark, a new benchmark for human-to-humanoid generalizable interaction skill transfer, encompassing 6 humanoid-scene interaction tasks spanning 11.8K diverse object shapes and 25K human references. In this section, we first introduce the simulation environment configurations of \benchmark in Section~\ref{sec:simulation_environment}. We then provide formulations and evaluation metrics of our six humanoid-scene interaction tasks in Section~\ref{sec:task_formulations}. To learn from human data, we finally present methods for human skill reference collection for each task in Section~\ref{sec:human_reference_collection}.

We contribute \benchmark, a new benchmark for human-to-humanoid generalizable interaction skill transfer, encompassing 6 humanoid-scene interaction tasks spanning 11.8K diverse object shapes and 25K human references. We first introduce the simulation environment configurations of \benchmark in Section~\ref{sec:simulation_environment}. We then provide formulations and evaluation metrics for our six humanoid-scene interaction tasks in Section~\ref{sec:task_formulations}. Finally, to facilitate learning from human data, we present the methods for human skill reference collection for each task in Section~\ref{sec:human_reference_collection}.

\subsection{Simulation Environment}
\label{sec:simulation_environment}

% As shown in Figure~\ref{fig:human_model_and_scene_layout} (a), \benchmark is constructed in Isaac Gym~\cite{makoviychuk2021isaac} simulation environment. The humanoid robot is Unitree H1~\cite{unitree_h1} consisting of 19 revolute joints. The physical model of H1 follows the Unitree public implementation~\cite{unitree_h1_codebase} for Isaac Gym. We describe key designs as follows and details in the supplementary material.

\begin{figure}[h!]
   \centering
   \includegraphics[width=1.0\linewidth]{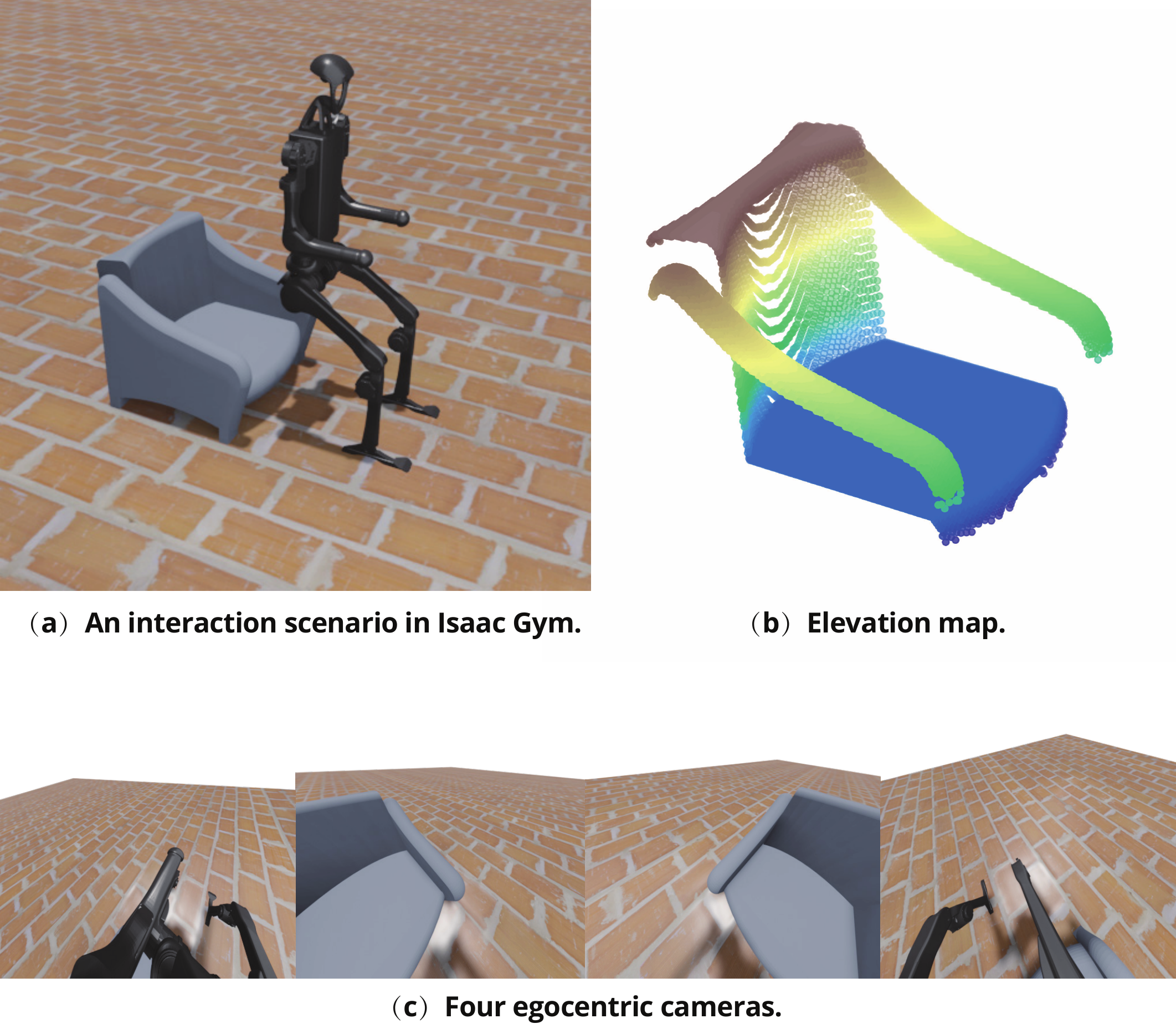}
   \vspace{-0.7cm}
   \caption{\textbf{\benchmark simulation configurations.} (a) exemplifies an interaction scenario of H1 in Isaac Gym. (b) and (c) show the captured elevation map and color images from four egocentric cameras.
   % ~\eric{The elevation map is strange. should be a 2D image no?}
   }
   \vspace{-15pt}
   \label{fig:human_model_and_scene_layout}
\end{figure}

As shown in Figure~\ref{fig:human_model_and_scene_layout} (a), \benchmark is constructed in the Isaac Gym~\cite{makoviychuk2021isaac} simulation environment. The humanoid robot used in the benchmark is the Unitree H1~\cite{unitree_h1}, consisting of 19 revolute joints. The physical model of the H1 follows the Unitree public implementation~\cite{unitree_h1_codebase} for Isaac Gym. Below, we describe the key design elements, with additional details in the supplementary.

% proprioception, ego camera setup, elevation map, figure 2 c,d.
\textbf{Observations.}
% \modify{vision and ego is important for real-world robot}.
% We draw inspiration from recent real-world vision-based humanoid control progresses~\cite{fu2024humanplus,he2024omnih2o} and build up an egocentric visual perception system for the humanoid. To obtain information from all directions around the humanoid body, we attach four egocentric cameras to the humanoid's root link, with a 120-degree field of view for each shown in Figure~\ref{fig:human_model_and_scene_layout} (c), capturing panoramic information and supporting humanoids to interact with objects from any direction. The \textbf{visual observations} of the humanoid include color $\mathcal{C}$, depth $\mathcal{D}$, and semantic segmentation $\mathcal{S}$ streams from the four cameras, combined with the elevation map~\cite{miki2022elevation} $\mathcal{E}$ shown in Figure~\ref{fig:human_model_and_scene_layout} (b) that is computed by merging depth images to the humanoid's root coordinate system and ground-projection. Besides, the humanoid \textbf{proprioception} $s_{\text{prop}}$ comprises 19 DoF joint angles, 19 DoF joint velocities, and 2 DoF gravity direction at the root of the humanoid. Overall, the complete observation space is $\mathcal{O} = \{s_{\text{prop}}, \mathcal{C}, \mathcal{D}, \mathcal{S}, \mathcal{E}\}$.
We draw inspiration from recent advancements in real-world vision-based humanoid control~\cite{fu2024humanplus,he2024omnih2o} and develop an egocentric visual perception system for the humanoid. To gather information from all directions around the humanoid body, we attach four egocentric cameras to the humanoid's root link, each with a 120-degree field of view, as shown in Figure~\ref{fig:human_model_and_scene_layout} (c). These cameras provide panoramic coverage, enabling the humanoid to interact with objects from any direction. The \textbf{visual observations} of the humanoid include color images $\mathcal{C}$, depth images $\mathcal{D}$, and semantic segmentation maps $\mathcal{S}$ from the four cameras, combined with an elevation map $\mathcal{E}$~\cite{miki2022elevation,jenelten2024dtc} shown in Figure~\ref{fig:human_model_and_scene_layout} (b). The elevation map is generated by merging depth images into the humanoid's root coordinate system and projecting them onto the ground. In addition, the humanoid’s \textbf{proprioception} $s_{\text{prop}}$ consists of 19 degrees of freedom (DoF) joint angles, 19 DoF joint velocities, and 2 DoF for the gravity direction at the humanoid's root. Combining all the sensory inputs, the complete observation space is defined as $\mathcal{O} = \{s_{\text{prop}}, \mathcal{C}, \mathcal{D}, \mathcal{S}, \mathcal{E}\}$.

% position control. we also support velocity and torque controls
\textbf{Actions.} \benchmark supports position, velocity, and torque control, all of which operate in 19-dimensional action spaces with a control frequency of 50 Hz. For both position and velocity controls, we employ a PD controller, using the same proportional-derivative gains as those in HumanPlus~\cite{fu2024humanplus}. In our experiments, we use position control due to its universality and to ensure consistency with existing real-world implementations~\cite{he2024omnih2o,fu2024humanplus,zhang2024wococo}.
 
\subsection{Task Formulations and Evaluations}
\label{sec:task_formulations}

Aiming to bridge the gap between graphical human interaction synthesis methods~\cite{starke2019neural,zhang2022couch,roam,unihsi,jiang2024scaling} and robotic humanoid skill learning~\cite{al2023locomujoco,sferrazza2024humanoidbench,chernyadev2024bigym}, \benchmark focuses on everyday household human-scene interaction scenarios that are widely studied in the computer vision community. This includes five tasks involving interactions with static scenes and one with dynamic scenes.

% \textbf{Evaluation metrics.} To evaluate whether methods can achieve the tasks with the potential of adapting to real-world robots, we propose using a combination of kinematic and physical metrics. The kinematic metric $K$ selects task-specific key joints of the humanoid and assesses whether they achieve pre-defined task goals, meanwhile, the physical metric $P=[\max(\Vert \tau_i v_i \Vert^2) < \lambda_P]$ ensures humanoids keep moving with small enough energy costs, where $i$, $\tau$, and $v$ denotes the frame index, mean joint torque, and mean joint velocity, respectively. Due to unknown expected values of $\lambda_P$ in future sim-to-real transfer, we currently select four values with exponential changes and report the average success rates evaluated under them.

\textbf{Evaluation metrics.} To assess whether methods can successfully achieve the tasks and adapt to real-world robots, we propose a combination of kinematic and physical metrics. The kinematic metric $K$ evaluates task-specific key joints of the humanoid to determine whether they meet predefined task goals. The physical metric $P=[\max(\Vert \tau_i v_i \Vert^2) < \lambda_P]$ ensures that humanoids perform movements with sufficiently low energy costs, where $i$, $\tau$, and $v$ denote the frame index, mean joint torque, and mean joint velocity, respectively. Since the expected values of $\lambda_P$ for sim-to-real transfer are unknown, we select four values with exponential increments and report the average success rates under these values.

We present brief formulations and kinematic evaluation metrics for each task below. Detailed formulations are provided in the appendix.

% detailed definitions of 6 tasks: input, output, evaluation metric
\textbf{(1) Sitting on a Chair (SC).} The humanoid must navigate the scene, locate, and stably sit on a fixed unseen chair. The kinematic evaluation ensures that the pelvis of the humanoid remains above the seat surface.

% \textbf{(2) Sitting on a sofa (SS).} The humanoid is required to find and steadily sit on a fixed unseen rigid sofa. Compared to chairs, sofas in our dataset comprise significantly longer lengths and shorter heights that induce different sitting poses of the humanoid and release its reliance on seat backs. The kinematic evaluation is the same as that for SC.%~\eric{need to point out why this is a different task from 1}
\textbf{(2) Sitting on a Sofa (SS).} The humanoid is required to find and sit steadily on a fixed unseen sofa. Compared to chairs, sofas in our dataset have significantly longer lengths and shorter heights, which lead to different sitting poses and reduce reliance on the seat back. The kinematic evaluation for this task is the same as that for SC.

% \textbf{(3) Lying on a bed (LB).} The humanoid needs to locate, sit, and finally lie on a fixed unseen bed. The kinematic evaluation constrains the pelvis and ankles of the humanoid to be kept on top of the bed.
\textbf{(3) Lying on a Bed (LB).} The humanoid needs to locate, sit, and finally lie on a fixed unseen bed. The kinematic evaluation ensures that both the pelvis and ankles of the humanoid remain on top of the bed.

% \textbf{(4) Lying on a sofa (LS).} The humanoid needs to locate, sit, and lie on a fixed unseen sofa. Since the humanoid cannot move the whole body onto sofas due to its large size, the kinematic evaluation, instead, requires the pelvis to be kept in the sofa, meanwhile requiring raising ankles to half of the sofa's height to ensure supports are only from the sofa.
\textbf{(4) Lying on a Sofa (LS).} The humanoid must locate, sit, and lie on a fixed unseen sofa. Since the humanoid is too large to move its entire body onto the sofa, the kinematic evaluation requires the pelvis to remain on the sofa, while the ankles should be raised to about half the sofa’s height to ensure that support is solely from the sofa.

% \textbf{(5) Touching points near an object (T).} With an unseen household object in the scene, the humanoid needs to move its wrists to target positions that are easy to manipulate the object. Target wrist positions are specified via markers on the elevation map as additional vision input. The kinematic evaluation requires wrists to be kept near target positions.
\textbf{(5) Touching Points Near an Object (T).} In this task, with an unseen household object in the scene, the humanoid must move its wrists to target positions that allow easy manipulation of the object. The target wrist positions are specified by markers on the elevation map, which is used as additional visual input. The kinematic evaluation ensures that the wrists remain near the target positions.
% ~\eric{need to modify the task description when vision is incorporated}

% \textbf{(6) Lifting a box (L).} Given a box with an unseen scale, the humanoid is required to locate and lift it to reach a specific height. The kinematic evaluation measures whether the box becomes above the target height at the end.
\textbf{(6) Lifting a Box (L).} Given a box of an unseen scale, the humanoid must locate and lift the box to reach a specific height. The kinematic evaluation measures whether the box is above the target height and the hands are attached to the object by the end of the task.

\subsection{Human Skill Reference Collection}
\label{sec:human_reference_collection}

% \modify{motivation, existing methods are teleoperation, why we use synthetic data}. 
Benefiting from advances in understanding and synthesizing human-scene interactions, we have integrated a large-scale human skill reference dataset for the six tasks. The data collection process for each task is described below.

% \textbf{Generating human-scene interaction by UniHSI~\cite{unihsi}.} For tasks sitting on chairs and lying on beds, we generate reference motions using UniHSI~\cite{unihsi}, which takes Chains of Contacts as input and outputs corresponding humanoid motions for the AMP~\cite{peng2021amp} robot. We utilize the policy trained on select scenes from PartNet~\cite{partnet} with corresponding Chains of Contacts. To scale data generation, we apply this process across all available scenes in PartNet~\cite{partnet} and manually generate Chains of Contacts for inference. For each motion, we enhance geometric diversity by randomly scaling and stretching objects as data augmentation, and we also randomize the robot's initial position. Finally, we filter all generated data to retain only successful interactions.

\textbf{Generating human-scene interaction by UniHSI~\cite{unihsi}.} For sitting on chairs and lying on beds, we generate reference motions using UniHSI~\cite{unihsi}, which takes Chains of Contacts as input and outputs humanoid motions for the AMP~\cite{peng2021amp} robot. We apply the policy trained on PartNet~\cite{partnet} scenes, manually generating Chains of Contacts for inference. To enhance diversity, we randomly scale and stretch objects, and vary the robot's initial position. Only successful interactions are retained.

% \textbf{Generating human-sofa interaction by ROAM~\cite{roam}.} For tasks sitting and lying on sofas, we use ROAM~\cite{roam} to generate reference motions in two stages. In the first stage, the goal pose synthesis network generates final sitting poses and matches them with object meshes (we select meshes from ShapeNet~\cite{shapenet}). Each mesh is randomly assigned multiple sitting poses, and to increase geometric diversity, we apply random scaling and stretching of the objects. In the second stage, ROAM generates the full motion sequence for each (mesh, final pose) pair, simulating the human walking toward the sofa and sitting down. The human character's initial position is randomly placed within a sector in front of the mesh, and its orientation is set to a random angle.

\textbf{Generating human-sofa interaction by ROAM~\cite{roam}.} For sitting and lying on sofas, we use ROAM~\cite{roam} in two stages. First, the goal pose synthesis network generates final sitting poses and matches them to object meshes (from ShapeNet~\cite{shapenet}). Multiple sitting poses are assigned to each mesh, with random scaling for diversity. In the second stage, ROAM generates the full motion sequence, simulating a human walking toward the sofa and sitting down. The human's initial position and orientation are randomly set.

% \textbf{Selecting interaction clips from CORE4D~\cite{core4d}.} For tasks touching points and lifting boxes, we acquire human skill references from an existing motion capture dataset CORE4D~\cite{core4d}. For task touching points, we automatically detect the frames where humans are about to touch the objects by hand-object distances and clip the interactions before these frames as human skill references. For task lifting boxes, we select 30 large boxes and clip corresponding interaction data by whether the box reaches the target height.
\textbf{Selecting interaction clips from CORE4D~\cite{core4d}.} For the tasks of touching points and lifting boxes, we acquire human skill references from the existing motion capture dataset CORE4D~\cite{core4d}. For the touching points task, we automatically detect the frames where humans are about to touch an object by calculating the hand-object distance and clip the interaction sequences leading up to these frames as human skill references. For the lifting boxes task, we select 30 large boxes and clip the corresponding interaction data based on whether the box reaches the target height.

\textbf{Dataset statistics.}  
Table \ref{tab:dataset_statistics} presents statistics on the dataset, showcasing a wide variety of objects and skill references, with significant diversity in object scales. By leveraging this dataset, \benchmark fosters studies on the generalizability of methods to changes in object geometry.

\begin{table}[h!]
\centering
\scriptsize
\addtolength{\tabcolsep}{-3pt}
{
\begin{tabular}{|c|cccc|cc|}
\hline
\multirow{2}{*}{Task} & \multicolumn{4}{c|}{Object} & \multicolumn{2}{c|}{Human Reference} \\
\cline{2-7}
& Number & Length (cm) & Width (cm) & Height (cm) & Number & Duration (s) \\
\hline
SC & 4.6K & 57.8($\pm$14.9) & 55.6($\pm$15.1) & 80.6($\pm$17.6) & 5.0K & 14.8($\pm$0.5) \\
SS & 1.5K & 131.1($\pm$21.1) & 63.0($\pm$21.7) & 57.3($\pm$21.8) & 7.6K & 12.1($\pm$0.0) \\
LB & 350 & 227.3($\pm$42.7) & 177.7($\pm$39.4) & 85.7($\pm$27.0) & 1.6K & 24.8($\pm$0.7) \\
LS & 1.8K & 191.2($\pm$28.3) & 86.7($\pm$25.2) & 78.5($\pm$25.7) & 6.1K & 12.1($\pm$0.0) \\
T & 2.3K & 59.6($\pm$19.2) & 53.2($\pm$16.2) & 56.1($\pm$16.7) & 2.3K & 3.3($\pm$0.2) \\
L & 30 & 67.3($\pm$6.4) & 43.4($\pm$6.2) & 88.9($\pm$7.8) & 890 & 4.3($\pm$0.2) \\
\hline
\end{tabular}
}
\vspace{-0.3cm}
\caption{\textbf{Statistics on human skill references.}}
\vspace{-0.5cm}
\label{tab:dataset_statistics}
\end{table}

\section{Skill Learning Paradigm}
\label{sec:skill_learning_paradigm}

\begin{figure*}[h!]
   \centering
   \includegraphics[width=0.9\textwidth]{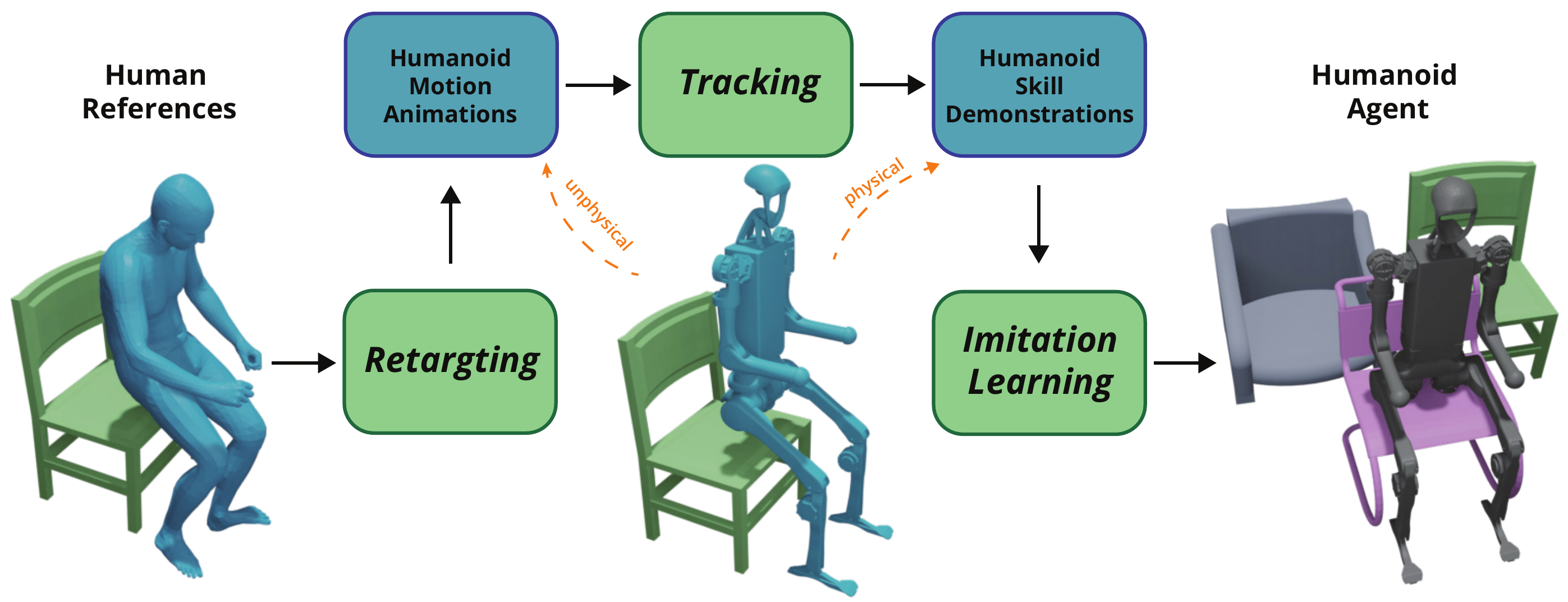}
   \vspace{-0.2cm}
   \caption{\textbf{Humanoid interaction skill learning paradigm.}}
   \vspace{-0.4cm}
   \label{fig:skill_learning_pipeline}
\end{figure*}

Learning humanoid skills from human data is an emerging field~\cite{phc,hassan2023synthesizing,fu2024humanplus,he2024omnih2o}, supported by large-scale motion datasets~\cite{mahmood2019amass} and manual teleoperation~\cite{ze2024generalizable}. A three-stage skill learning paradigm, shown in Figure~\ref{fig:skill_learning_pipeline}, integrates key techniques from recent work~\cite{he2024omnih2o,fu2024humanplus,tessler2024maskedmimic}. First, a \textbf{retargeting module}~\cite{tang2024humanmimic,he2024omnih2o} (Section~\ref{sec:retargeting}) transfers human motions to humanoid models to generate motion animations. Next, a \textbf{motion tracking module}~\cite{phc,fu2024humanplus,tessler2024maskedmimic} (Section~\ref{sec:tracking}) is trained in simulations to execute realistic motions based on these animations, producing skill demonstrations. Finally, an \textbf{imitation learning module}~\cite{act,diffusion_policy,pulse} (Section~\ref{sec:IL}) learns from these demonstrations to train humanoid agents for unseen scenarios. Our \benchmark implements this paradigm for six tasks and benchmarks advanced methods for each module, enabling comprehensive evaluation.

\subsection{Humanoid Motion Retargeting}
\label{sec:retargeting}

% To build up the bridge between human skill references and humanoid animations, the motion retargeting module transfers motions
% % from a human skeleton $\mathcal{A}$=$\{R_{\mathcal{A}} \in \mathbb{R}^{T\times3}, t_{\mathcal{A}} \in \mathbb{R}^{T\times3}, \theta_{\mathcal{A}} \in \mathbb{R}^{T\times N_\mathcal{A}}\}$ to a humanoid skeleton $\mathcal{B}$=$\{R_{\mathcal{B}} \in \mathbb{R}^{T\times3}, t_{\mathcal{B}} \in \mathbb{R}^{T\times3}, \theta_{\mathcal{B}} \in \mathbb{R}^{T\times N_\mathcal{B}}\}$, where $R$, $t$, $\theta$ denotes the root orientation, root position, and joint angles of the skeleton in each frame. $T$ and $N$ denote the frame and joint numbers, respectively.
% from a human skeleton $\mathcal{A}$ to a humanoid skeleton $\mathcal{B}$.
% To correlate $\mathcal{A}$ and $\mathcal{B}$, existing retargeting approaches select key joints from both skeletons and manually define a joint mapping $\mathcal{M}$=$\{<a_k, b_k>\}$ \cite{tang2024humanmimic}, and then diverges into three types with different retargeting goals. We describe each method type below and incorporate them all in \benchmark.

To bridge human skill references with humanoid animations, the motion retargeting module transfers motions from a human skeleton $\mathcal{A}$ to a humanoid skeleton $\mathcal{B}$. To establish a correlation between $\mathcal{A}$ and $\mathcal{B}$, existing retargeting approaches typically select key joints from both skeletons and manually define a joint mapping $\mathcal{M} = \{<a_k, b_k>\}$~\cite{tang2024humanmimic}.  
These approaches then diverge into three types, each with distinct retargeting goals. 
% We describe each method type below and incorporate them into \benchmark.

% (1) \textbf{Copy-rotation:} Given a predefined mapping $\mathcal{M}$, a straightforward strategy~\cite{fu2024humanplus} is to copy the joint angles from $a_k$ to $b_k$, directly. With the same definitions of root for $\mathcal{A}$ and $\mathcal{B}$, the root positions and orientations of $\mathcal{B}$ are also copied from those of $\mathcal{A}$. This method acquires realistic joint orientations without ensuring joint position correctness.
(1) \textbf{Copy-rotation:} Given a predefined joint mapping $\mathcal{M}$, a straightforward strategy~\cite{fu2024humanplus} is to directly copy the joint angles from $a_k$ to $b_k$. In this method, the root positions and orientations of $\mathcal{B}$ are also copied from those of $\mathcal{A}$, assuming both skeletons share the same root definitions. This approach results in realistic joint orientations but does not guarantee the correctness of joint positions.

% (2) \textbf{Optimization:} Contrary to copy-rotation, the optimization method~\cite{radosavovic2024humanoid,jiang2024harmon} aims at recovering precise joint positions while releasing joint angle fidelity by maximizing specific energy functions. Energy functions commonly involve constraints for joint global positions, global orientations, and ranges of joint angles.
(2) \textbf{Optimization:} In contrast to copy-rotation, the optimization method~\cite{radosavovic2024humanoid, jiang2024harmon} focuses on recovering accurate joint positions while allowing for some loss in joint angle fidelity. The optimization is achieved by maximizing energy functions that incorporate constraints on joint global positions, global orientations, and joint angle ranges.

% (3) \textbf{Optimization with skeleton shape alignment:} To find a trade-off between joint angle and global position fidelities, recent methods~\cite{he2024omnih2o,tang2024humanmimic,he2024learning} propose to first adjust bone lengths of $\mathcal{A}$ to map those of $\mathcal{B}$ and then conduct aforementioned optimization. The adjustment for $\mathcal{A}$ maintains joint angles and changes joint positions, while the following optimization is the opposite.
(3) \textbf{Optimization with skeleton shape alignment:} To strike a balance between joint angle fidelity and global position accuracy, recent methods~\cite{he2024omnih2o, tang2024humanmimic, he2024learning} propose a two-step process. First, they adjust the bone lengths of $\mathcal{A}$ to align with those of $\mathcal{B}$, and then perform optimization as described above. This adjustment alters joint positions while maintaining joint angles, while the subsequent optimization process aims to refine the joint angles.

For each human skeleton, we implement the corresponding copy-rotation and optimization algorithms, and use OmniH2O~\cite{he2024omnih2o} to represent the optimization with skeleton shape alignment. The kinematic metric is applied to filter the obtained animations, passing only the successful ones to the next module.

\subsection{Humanoid Motion Tracking}
\label{sec:tracking}

Regardless of whether the human reference motions are physical, the retargeted humanoid motion $\mathcal{B}$ (as described in Section~\ref{sec:retargeting}) is unphysical. To generate physical humanoid skill demonstrations and corresponding action sequences of the animation, we apply a tracking module to $\mathcal{B}$. For the tracking module, we benchmark HST~\cite{fu2024humanplus} and a simplified version of PHC~\cite{phc}. We briefly describe the key features and differences between these two trackers, as well as introduce the design of task-specific reward functions.

\textbf{Simplified PHC~\cite{phc}.} PHC~\cite{phc} formulates a Markov Decesion Process (MDP) policy $\pi$ that is trained using proximal policy optimization (PPO)~\cite{ppo}. At each simulation step, the policy $\pi$ takes humanoid's proprioception and goal state as inputs, and outputs the humanoid action to be performed at the next step. A proportional-derivative (PD) controller is applied to each degree of freedom (DoF) of the humanoid, with the action specifying the PD target. Reference state initialization\cite{peng2018deepmimic} is used during training, along with early termination to stop the process when the tracking error exceeds a threshold. To simplify the training process, we use a single PHC policy, omitting the suggested composer~\cite{phc} that combines multiple policies.

\textbf{Improved HST~\cite{fu2024humanplus}.} Similar to PHC, HST uses PPO~\cite{ppo} to learn a policy generating humanoid actions based on proprioception and goal state inputs. In practice, we improve the official HST implementation~\cite{humanplus_codebase} with redundant proprioception representations from MaskedMimic~\cite{tessler2024maskedmimic}, early-termination strategies inspired by DeepMimic~\cite{peng2018deepmimic}, and motion regularization rewards for stable policy training~\cite{he2024omnih2o,wang2024skillmimic,portela2024learning}. The main difference between the two trackers is that the simplified PHC incorporates a higher-dimensional goal state input (primarily including state differences), whereas our improved HST includes additional reward constraints on motion energy.

\textbf{Task-specific reward functions.} For tasks SC, SS, LB, and LS, the primary tracking reward aims to minimize joint position and rotation angle errors between the humanoid and the reference motion $\mathcal{B}$. For tasks T and L, greater emphasis is placed on the motion of the hands relative to the object (for task L, this also includes the object itself). Inspired by PhysHOI~\cite{wang2023physhoi}, we design a reward function that combines joint position, root alignment, and hand position rewards, with a higher weight assigned to the hand position. For task L, additional focus is given to the reward for object positioning and the relative relationship between the hand and the object. Details are in the appendix.

% After training the above motion trackers, we use them to track the animations from their training sets and acquire humanoid skill demonstrations. We then filter the demonstrations with the kinematic metric and input successful ones to the next module.
After training the motion trackers described above, we use them to track the animations from their respective training sets and obtain humanoid skill demonstrations. These demonstrations are then filtered using the kinematic metric, and only the successful ones are passed to the next module.
\vspace{-15pt}

\subsection{Humanoid Imitation Learning}
\label{sec:IL}
\vspace{-5pt}

% As introduced in Section~\ref{sec:related_work_IL}, the imitation learning module aims to train a generic humanoid agent using humanoid skill demonstrations, where the agent needs to fulfill the tasks in unseen scenarios. We select ACT~\cite{fu2024humanplus} and HIT~\cite{fu2024humanplus} as representatives of the single-stage and two-stage algorithms and incorporate them into our experiments. We briefly describe the two methods below and provide details in supplementary material.
As introduced in Section~\ref{sec:related_work_IL}, the imitation learning module aims to train a generic humanoid agent using humanoid skill demonstrations, with the goal of enabling the agent to perform tasks in unseen scenarios. In our experiments, we select ACT~\cite{fu2024humanplus} and HIT~\cite{fu2024humanplus} as representatives of the single-stage and two-stage algorithms, respectively. 
% Below, we briefly describe the two methods, with more details provided in the supplementary material.

% \textbf{ACT~\cite{act}:} ACT is a single-stage method that generates the robot's action sequences directly from vision and proprioception inputs, consisting of a CVAE-based~\cite{CVAE} Transformer encoder-decoder structure. ACT has shown superior performances on existing robot learning scenes~\cite{shi2023waypoint} and benchmarks~\cite{chernyadev2024bigym,jia2024towards}.
\textbf{ACT~\cite{act}:} ACT is a single-stage method that generates the robot's action sequences directly from vision and proprioception inputs. It uses a CVAE-based~\cite{CVAE} Transformer encoder-decoder structure. ACT has demonstrated superior performance on existing robot learning tasks~\cite{shi2023waypoint} and benchmarks~\cite{chernyadev2024bigym,jia2024towards}.

% \textbf{HIT~\cite{fu2024humanplus}:} HST is a two-stage approach that integrates a high-level Transformer-based motion planner generating real-time robot animations and a low-level motion tracker physically mimicking the animations. This method can be easily incorporated into our skill-learning paradigm with pre-trained trackers in Section~\ref{sec:tracking}.
\textbf{HIT~\cite{fu2024humanplus}:} HIT is a two-stage approach that integrates a high-level, Transformer-based motion planner, which generates real-time robot animations, and a low-level motion tracker that physically mimics the animations. This method can be easily incorporated into our skill-learning paradigm by using pre-trained trackers, as described in Section~\ref{sec:tracking}.
\vspace{-5pt}

% add GAIL - retargeting - IL, no tracking, make the paradigm abundant

\section{Experiments}
\vspace{-5pt}
Based on the multi-stage skill learning paradigm, \benchmark supports evaluations on both integrated learning pipelines~\cite{he2024omnih2o,fu2024humanplus} and modular learning algorithms~\cite{phc,act}.
We conduct extensive experiments across diverse combinations of visual input modalities (Section~\ref{sec:simulation_environment}) and algorithm choices for different modules (Section~\ref{sec:skill_learning_paradigm}), selecting the best combination as our current solution for the benchmark. This solution is then compared with existing pipelines (Section~\ref{sec:experiment_pipeline}) and ablated across various modules (Sections~\ref{sec:experiment_retargeting}, \ref{sec:experiment_tracking}, \ref{sec:experiment_IL}), visual modalities (Section~\ref{sec:experiment_vision}), and human reference scales (Section~\ref{sec:experiment_data_scale}).

\begin{figure}[ht!]
    \centering
    \includegraphics[width=\linewidth]{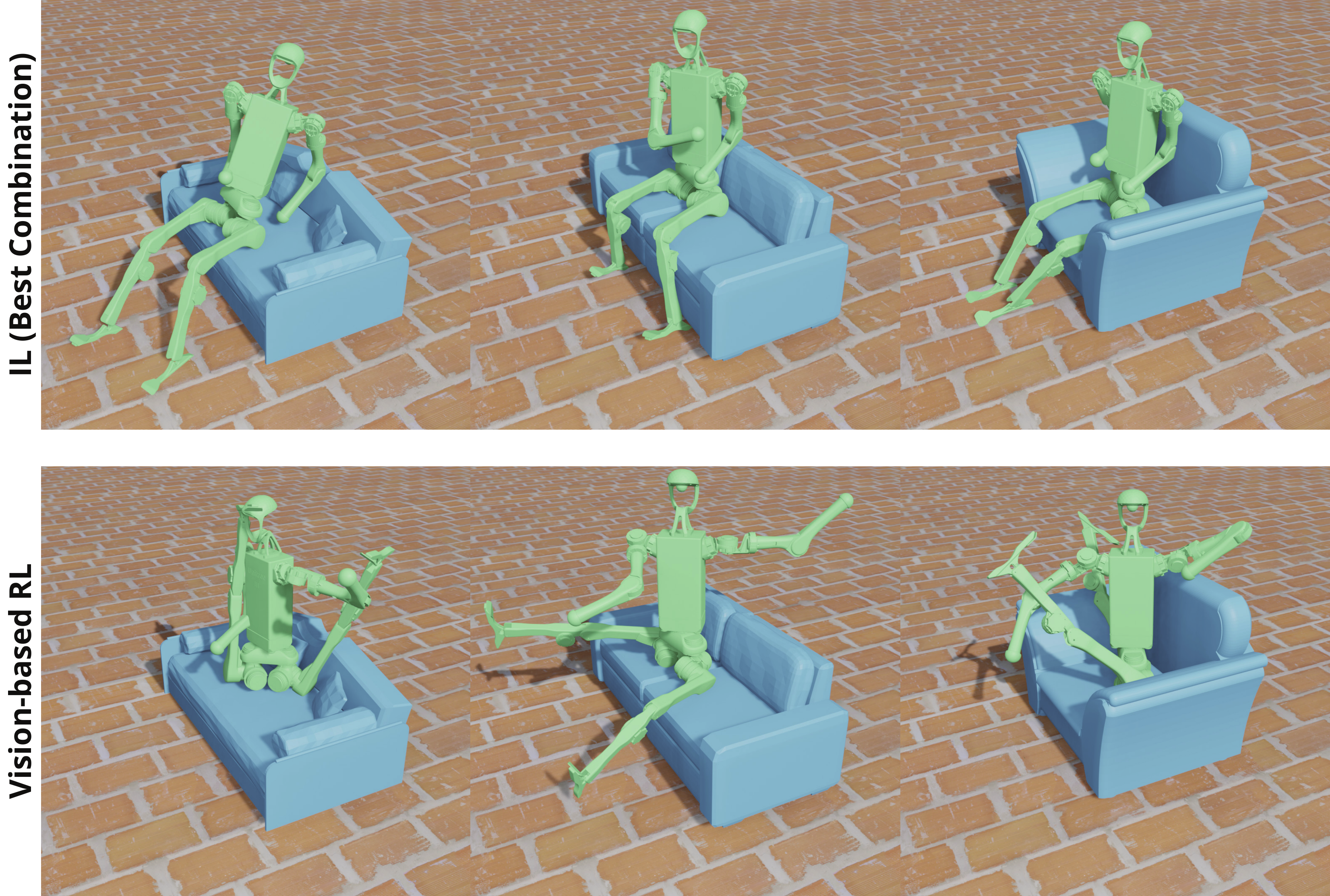}
    \vspace{-0.6cm}
    \caption{\textbf{Qualitative comparisons on data-driven human mimicking and data-free RL on sitting sofas.} RL struggles to get reasonable poses despite completing the task kinematically.}
    \vspace{-0.2cm}
    \label{fig:ILRL_compare}
\end{figure}

% \textbf{The best combination} is leveraging elevation map $\mathcal{E}$ alone as visual observation and using the optimization, improved HST~\cite{fu2024humanplus}, and ACT~\cite{act} in the retargeting, tracking, and imitation learning modules, respectively. We anticipate this solution to be a baseline method in future research.
\textbf{The best combination} involves using the elevation map $\mathcal{E}$ as the sole visual input, and employing optimization, improved HST~\cite{fu2024humanplus}, and ACT~\cite{act} for the retargeting, tracking, and imitation learning modules, respectively. We consider this solution a baseline for future research in this area.

% \textbf{Train-test splits:} For each task, we randomly split objects in the dataset into training and test sets following a ratio of 3:1. We use training sets only to train modular algorithms, and deploy trained humanoid agents on test sets to assess their generalizability on unseen objects.
\textbf{Train-test splits:} For each task, we randomly divide the objects in the dataset into training and test sets, following a $3:1$ ratio. The training sets are used exclusively to train the modular algorithms, while the trained humanoid agents are deployed on the test sets to evaluate their generalizability.% on unseen objects.

% \textbf{Notations for evaluation results:} In the following tables, we evaluate the kinematic success rate determined by the kinematic metric ($K$) itself, and the energy-averaged success rate judged by $K$ and the physical metric $P$ with four different $\lambda_P$ described in Section~\ref{sec:task_formulations}. We report the two types of success rates separated by a forward slash.
\textbf{Notations for evaluation results:} In the following tables, we report both the kinematic success rate, determined by the kinematic metric ($K$), and the energy-averaged success rate, which is judged by $K$ and the physical metric $P$ using four different $\lambda_P$ values, as described in Section~\ref{sec:task_formulations}. These two success rates are separated by a forward slash.

\subsection{Skill Learning Pipelines}
\label{sec:experiment_pipeline}

We compare the best combination of our methods with two existing skill learning pipelines: OmniH2O~\cite{he2024omnih2o} and HumanPlus~\cite{fu2024humanplus}. OmniH2O provides retargeting and tracking, to which we add our best imitation learning algorithm. HumanPlus offers an end-to-end pipeline with a novel imitation learning algorithm. Additionally, we evaluate a vision-based reinforcement learning (RL) method~\cite{ppo}, which learns entirely through environment exploration without human references. Table \ref{tab:pipeline_comparison} shows task success rates for these methods. Compared to the data-driven pipelines, vision-based RL~\cite{ppo} performs significantly worse, especially in terms of physical metrics, highlighting its poor physical plausibility, as shown in Figure~\ref{fig:ILRL_compare}. This emphasizes the importance of learning by mimicking human references.

% As shown in Figure~\ref{fig:ILRL_compare}, RL is less physically plausible because its only focus on task success, and it's hard to explore reasonable poses without humanoid skill demonstrations. In contrast, the best combination not only has a higher success rate, but also has a more reasonable humanoid posture which confirms the effectiveness of imitation learning from demonstration data.
Existing methodologies consistently face significant challenges on tasks touching points and lifting boxes due to difficulties in stably controlling hands and manipulating dynamic objects, revealing new research opportunities.

\begin{table}[h!]
\centering
\footnotesize
\addtolength{\tabcolsep}{-3pt}
{
\begin{tabular}{|c|c|cccccc|c|}
\hline
Method & Data & SC & SS & LB & LS & T & L & Mean \\
\hline
% Vision-based PPO~\cite{ppo} & \XSolidBrush & 50.5/24.7 & 43.9/20.2 & 18.0/8.3 & 53.8/25.9 & 0.0/0.0 & 0.0/0.0 & 27.7/13.2 \\
PPO~\cite{ppo} & \XSolidBrush & 50/25 & 44/20 & 18/9 & 54/26 & 0/0 & 0/0 & 28/13 \\
\hline
% state-based GAIL & \checkmark & & & & & & & \\
HumanPlus~\cite{fu2024humanplus} & \checkmark & 36/28 & 17/16 & 36/34 & 38/36 & \textbf{46}/\textbf{32} & 4/3 & 30/25 \\
OmniH2O~\cite{he2024omnih2o} & \checkmark & 79/\textbf{72} & 69/69 & 48/47 & 47/46 & 0/0 & 0/0 & 41/39 \\
Best Combination & \checkmark & \textbf{81}/69 & \textbf{84}/\textbf{84} & \textbf{61}/\textbf{61} & \textbf{80}/\textbf{78} & 11/10 & \textbf{6}/\textbf{5} & \textbf{54}/\textbf{51} \\
\hline
\end{tabular}
}
\vspace{-0.2cm}
\caption{Task success rates for different \textbf{skill learning pipelines}.}
\vspace{-0.4cm}
\label{tab:pipeline_comparison}
\end{table}

\subsection{Humanoid Motion Retargeting}
\label{sec:experiment_retargeting}

We evaluate the effect of combining the three motion retargeting algorithms with the best designs of other modules. Results are shown in Table \ref{tab:retargeting_comparison}. Studying retargeting algorithms is crucial for the success of the entire paradigm, as they directly impact the quality of humanoid motion and interaction. Among the different approaches, optimization consistently yields the best results across all settings, highlighting the importance of fine-grained positional accuracy for achieving natural and precise interactions. In tasks T and L, where hand position fidelity plays a critical role, only the optimization method can accurately capture the detailed hand movements relative to the objects, further underscoring the necessity of high precision in hand tracking.

\begin{figure*}[ht!]
    \centering
    \includegraphics[width=0.9\linewidth]{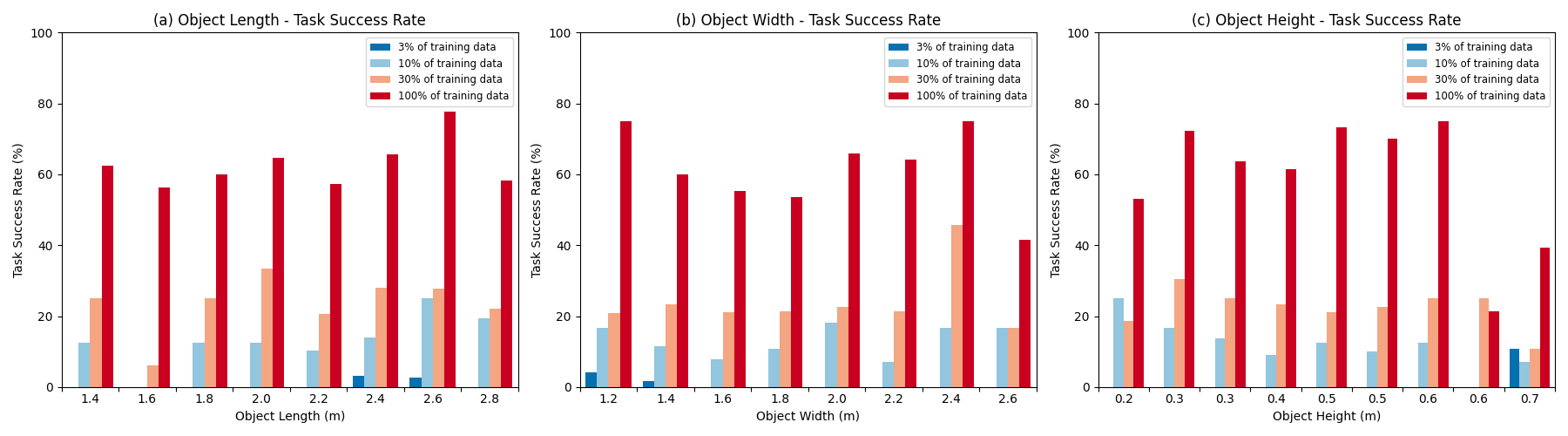}
    \vspace{-0.3cm}
    \caption{\textbf{LB task success rates on varying object sizes.} Object length/width refers to the size of bounding boxes and object height refers to the height of lying planes of beds.}
    \vspace{-10pt}
    \label{fig:liebed_scale}
\end{figure*}

\begin{table*}[ht!]
\centering
\footnotesize
\addtolength{\tabcolsep}{-3pt}
{
\begin{tabular}{|c|cccccc|c|cccccc|c|}
\hline
\multirow{2}{*}{Method} & \multicolumn{7}{c|}{Motion Tracking Performance} & \multicolumn{7}{c|}{Imitation Learning Performance} \\
\cline{2-15}
 & SC & SS & LB & LS & T & L & Mean & SC & SS & LB & LS & T & L & Mean \\
\hline
Copy-rotation & 66/46 & \textbf{90}/89 & 84/77 & 38/36 & 0/0 & 0/0 & 46/41 & 61/53 & \textbf{89}/\textbf{89} & 33/32 & 69/68 & 0/0 & 0/0 & 42/40 \\
OmniH2O \cite{he2024omnih2o} & 83/64 & \textbf{90}/\textbf{90} & 61/42 & 40/32 & 27/15 & 0/0 & 50/41 & 79/\textbf{72} & 69/69 & 48/47 & 47/46 & 0/0 & 0/0 & 41/39 \\
Optimization & \textbf{92}/\textbf{70} & \textbf{90}/89 & \textbf{88}/\textbf{82} & \textbf{55}/\textbf{53} & \textbf{58}/\textbf{40} & \textbf{25}/\textbf{20} & \textbf{68}/\textbf{59} & \textbf{81}/69 & 84/84 & \textbf{61}/\textbf{61} & \textbf{80}/\textbf{78} & \textbf{11}/\textbf{10} & \textbf{6}/\textbf{5} & \textbf{54}/\textbf{51} \\
\hline
\end{tabular}
}
\vspace{-0.3cm}
\caption{Task success rates ($\%$) for different \textbf{motion retargeting algorithms}. Motion Tracking Performance denotes evaluation results on tracked humanoid skill demonstrations, while Imitation Learning Performance denotes evaluation results on humanoid agents.}
\vspace{-0.2cm}
\label{tab:retargeting_comparison}
\end{table*}

\begin{table*}[t!]
\centering
\footnotesize
\addtolength{\tabcolsep}{-3pt}
{
\begin{tabular}{|c|cccccc|c|cccccc|c|}
\hline
\multirow{2}{*}{Method} & \multicolumn{7}{c|}{Motion Tracking Performance} & \multicolumn{7}{c|}{Imitation Learning Performance} \\
\cline{2-15}
 & SC & SS & LB & LS & T & L & Mean & SC & SS & LB & LS & T & L & Mean \\
\hline
Simplified PHC \cite{phc} & 87/65 & \textbf{92}/83 & 63/49 & \textbf{80}/\textbf{61} & 8/6 & 0/0 & 55/44 & 44/35 & 69/64 & 14/13 & 71/51 & 0/0 & 0/0 & 33/27 \\
Improved HST \cite{fu2024humanplus} & \textbf{92}/\textbf{70} & 90/\textbf{89} & \textbf{88}/\textbf{82} & 55/53 & \textbf{58}/\textbf{40} & \textbf{25}/\textbf{20} & \textbf{68}/\textbf{59} & \textbf{81}/\textbf{69} & \textbf{84}/\textbf{84} & \textbf{61}/\textbf{61} & \textbf{80}/\textbf{78} & \textbf{11}/\textbf{10} & \textbf{6}/\textbf{5} & \textbf{54}/\textbf{51} \\
\hline
\end{tabular}
}
\vspace{-0.3cm}
\caption{Task success rates ($\%$) for different \textbf{motion tracking algorithms}.}
\vspace{-0.4cm}
\label{tab:tracking_comparison}
\end{table*}

\subsection{Humanoid Motion Tracking}
\label{sec:experiment_tracking}

% HST, simplified PHC, (MaskedMimic to supp)

% We ablate the humanoid motion tracking algorithm in Table \ref{tab:tracking_comparison}. Analyzing the results reveals that selecting an appropriate tracking algorithm is crucial for the success of the entire paradigm. For motion tracking performance, both Improved HST~\cite{fu2024humanplus} and Simplified PHC~\cite{phc} perform well on tasks SC, SS, LB, and LS. However, on tasks T and L, HST significantly outperforms PHC. The inclusion of energy rewards plays a key role in this performance, as they help prevent excessive and unrealistic energy consumption, which is essential given our metric’s maximum energy constraint. This optimization ensures that energy expenditure stays within reasonable limits, promoting more efficient and physically plausible behaviors. The superior performance of HST, especially in tasks where precise tracking is crucial, underscores the importance of regularizers for achieving accurate humanoid interactions.

We ablate the humanoid motion tracking algorithm in Table \ref{tab:tracking_comparison}. The results highlight that selecting the right tracking algorithm is critical for the success of the entire paradigm. Both Improved HST~\cite{fu2024humanplus} and Simplified PHC~\cite{phc} perform well on tasks SC, SS, LB, and LS. However, on tasks T and L, HST significantly outperforms PHC. The inclusion of energy rewards plays a key role, preventing excessive energy consumption and ensuring efficient, physically plausible behaviors within the maximum energy constraint. The superior performance of HST, particularly in tasks requiring precise tracking, underscores the importance of regularizers for accurate humanoid interactions.

\subsection{Humanoid Imitation Learning}
\label{sec:experiment_IL}

% HIT, ACT, (optional) GAIL in supp

% compare to the vision-based RL (expected to be nearly 0 accuracy) \licheng{todo: explain results of vision-based RL}(Table \ref{tab:vision_RL})

% Table \ref{tab:IL_comparison} compares performances of combining the two imitation learning methods with optimal retargeting and tracking algorithms. The large performance gap between HIT~\cite{fu2024humanplus} and ACT~\cite{act} reveals the significance of imitation learning algorithm designs for the skill learning paradigm. Compared to ACT, HIT causes performance drops on five tasks, which could caused by rapidly increasing cumulative errors that induce the humanoid states to exceed distributions of both the tracker and HIT.

Table \ref{tab:IL_comparison} compares the performance of combining two imitation learning methods with the optimal retargeting and tracking algorithms. The significant performance gap between HIT~\cite{fu2024humanplus} and ACT~\cite{act} underscores the importance of carefully designed imitation learning algorithms within the skill learning paradigm. Compared to ACT, HIT leads to performance drops on five tasks, which may be caused by rapidly accumulating errors that push the humanoid states beyond the distributions of both the tracker and HIT, resulting in degraded performance.

% \modify{analyze the results. key messages: 1) studying IL is important for the whole paradigm, 2) ACT is better than HST, the reason could be mutual out-of-other's-distribution of the tracker and IL method in HST (it's an assumption).}

\begin{table}[h!]
\centering
\footnotesize
\addtolength{\tabcolsep}{-3pt}
{
\begin{tabular}{|c|cccccc|c|}
\hline
Method & SC & SS & LB & LS & T & L & Mean \\
\hline
HIT \cite{fu2024humanplus} & 36/28 & 17/16 & 36/34 & 38/36 & \textbf{46}/\textbf{32} & 4/3 & 30/25 \\
ACT \cite{he2024omnih2o} & \textbf{81}/\textbf{69} & \textbf{84}/\textbf{84} & \textbf{61}/\textbf{61} & \textbf{80}/\textbf{78} & 11/10 & \textbf{6}/\textbf{5} & \textbf{54}/\textbf{51} \\
\hline
\end{tabular}
}
\vspace{-0.2cm}
\caption{Task success rates ($\%$) for \textbf{imitation learning methods}.}
\vspace{-0.4cm}
\label{tab:IL_comparison}
\end{table}

\subsection{Egocentric Visual Perception Modalities}
\label{sec:experiment_vision}

% elevation map, ego RGB

% \modify{move this subsection to before 5.1}

With versatile choices of egocentric visual observations, \benchmark promotes studies on exploring the most effective visual modalities. We compare two common designs for visual observation in Table~\ref{tab:vision_comparison}. Analyzing the results, we observe that the elevation map (\(\mathcal{E}\)) significantly outperforms the multi-view RGBD (\(\mathcal{C}+\mathcal{D}\)) across all tasks. One possible reason for this is the challenge of interpreting egocentric images that contain much irrelevant or noisy information, which can hinder the agent's ability to focus on task-relevant features. In contrast, the elevation map provides a more structured and focused representation of the environment, which likely contributes to its superior performance. Based on these findings, we propose using the elevation map as the agent's visual input and explore various technical designs under this setting.

\vspace{-5pt}

\begin{table}[h!]
\centering
\footnotesize
\addtolength{\tabcolsep}{-3pt}
{
\begin{tabular}{|c|cccccc|c|}
\hline
Visual Observation & SC & SS & LB & LS & T & L & Mean \\
\hline
$\mathcal{C}+\mathcal{D}$ & 0/0 & 0/0 & 1/1 & 13/13 & 8/6 & 1/1 & 4/4 \\
$\mathcal{E}$ & \textbf{81}/\textbf{69} & \textbf{84}/\textbf{84} & \textbf{61}/\textbf{61} & \textbf{80}/\textbf{78} & \textbf{11}/\textbf{10} & \textbf{6}/\textbf{5} & \textbf{54}/\textbf{51} \\
\hline
\end{tabular}
}
\vspace{-0.3cm}
\caption{Task success rates ($\%$) for different \textbf{vision modalities}.}
\vspace{-0.7cm}
\label{tab:vision_comparison}
\end{table}

\begin{table}[h!]
\centering
\footnotesize
\addtolength{\tabcolsep}{-3pt}
{
\begin{tabular}{|c|cccccc|c|}
\hline
Training Data Scale & SC & SS & LB & LS & T & L & Mean \\
\hline
10$\%$ & 69/\textbf{69} & 80/80 & 13/12 & 52/52 & 1/1 & 0/0 & 36/35 \\
100$\%$ & \textbf{81}/\textbf{69} & \textbf{84}/\textbf{84} & \textbf{61}/\textbf{61} & \textbf{80}/\textbf{78} & \textbf{11}/\textbf{10} & \textbf{6}/\textbf{5} & \textbf{54}/\textbf{51} \\
\hline
\end{tabular}
}
\vspace{-0.3cm}
\caption{Task success rates ($\%$) for different \textbf{training data scales}.}
\vspace{-0.4cm}
\label{tab:data_scale_comparison}
\end{table}

\subsection{Scale of Human Skill Data}
\label{sec:experiment_data_scale}
\vspace{-0.1cm}

We compare the results of the best algorithm combination under different scales of human references. The comparison between 10\% and 100\% training data in Table~\ref{tab:data_scale_comparison} highlights the impact of data scale on performance. As expected, a larger training dataset leads to improved task success across most metrics. Specifically, increasing the data scale from 10$\%$ to 100$\%$ significantly boosts performance in tasks such as SC, SS, LB, and LS. For LB task, we expand the data scale comparison among  3\%, 10\%, 30\% and 100\%. Results in Fig.~\ref{fig:liebed_scale} show a significant performance improvement as the data scale increases, particularly for objects of extreme sizes. This demonstrates that large-scale data is crucial for training robust models, confirming the usefulness of our large-scale dataset for improving generalizability and performance across various tasks.

% \licheng{analysis Figure~\ref{fig:data_scale_evaluation} if we had one.}

% \begin{figure}[h!]
%    \centering
%    \fbox{\rule{0pt}{2in} \rule{0.9\linewidth}{0pt}}
%    \caption{Task success rates using various scales of human references.}
%    \label{fig:data_scale_evaluation}
% \end{figure}

\section{Limitation and Conclusion}
\vspace{-5pt}
% We contribute \benchmark, the first comprehensive benchmark supporting generalizable humanoid-scene interaction skill learning via mimicking human references, which comprises six tasks and an integrated human reference dataset spanning 11K object shapes and 23K human-scene interaction motions. To advance research on various technical challenges, we construct a generic humanoid interaction skill learning paradigm with three key technical modules and present pipeline-wise and modular method comparisons. Extensive experiments demonstrate the effectiveness of learning from human mimicking, reveal the significance of modular algorithm selection, and pose new challenges and research chances to existing methodologies.
We introduce \benchmark, the first comprehensive benchmark designed to support generalizable humanoid-scene interaction skill learning through the mimicking of human references. It encompasses six diverse tasks and an integrated human reference dataset, featuring 11K object shapes and 23K human-scene interaction motions. To advance research on various technical challenges, we present a generic humanoid interaction skill learning paradigm comprising three key modules, alongside pipeline-wise and modular method comparisons. Extensive experiments demonstrate the effectiveness of learning from human imitation, highlight the importance of selecting appropriate algorithms for each module, and identify new challenges and research opportunities for existing methodologies.

% \textbf{Limitation.} Dexterous hands are not attached to our robot model since our focus is torso-level interaction. Incorporating hands could be an interesting future direction fostering research on find-grained manipulation skills.
\textbf{Limitation.} Our robot model does not include dexterous hands, as our focus is primarily on torso-level interaction. Future work could explore incorporating hands, which would enable the study of fine-grained manipulation skills.

% Secondly, articulated and soft objects are not covered in \benchmark due to the lack of relevant human datasets and interaction generation methods.

% citations
{
    \small
    \bibliographystyle{ieeenat_fullname}
    \bibliography{ref}
}

% WARNING: do not forget to delete the supplementary pages from your submission 
\clearpage
\Large \textbf{Appendix}
\\
\normalsize
\appendix
{\Large \textbf{Contents:}}
\begin{itemize}
\item \ref{supp_sec:additional_related_work}. Additional Related Work
\item \ref{supp_sec:experiments_tracking_IL}. Supplementary Experiments on Motion Tracking and Imitation Learning
\item \ref{supp_sec:simulation_enviroment}. Details on Simulation Environment Configurations
\item \ref{supp_sec:task_evaluation_metrics}. Details on \benchmark Task Evaluation Metrics
\item \ref{supp_sec:retargeting}. Details on Motion Retargeting Algorithms
\item \ref{supp_sec:tracking}. Motion Tracking Algorithm Designs
\item \ref{supp_sec:IL}. Imitation Learning Algorithm Designs
\item \ref{supp_sec:skill_learning_visualizations}. Skill Learning Visualizations
\end{itemize}

\section{Additional Related Work}
\label{supp_sec:additional_related_work}

\subsection{Human-to-Humanoid Skill Transfer}

Transferring skills from humans to humanoid robots is a popular research topic that guides humanoids to fulfill tasks in human-like manners. For locomotion skill learning, existing works~\cite{radosavovic2024humanoid,cheng2024expressive} integrate motion capture and synthetic human data as we do and adopt imitation learning on these data. To learn upper-body manipulation skills, previous studies~\cite{TRILL,li2024okami,jiang2024dexmimicgen,ze2024generalizable} propose learning from humanoid skill demonstrations collected by manual teleoperation due to the lack of relevant dataset support. Focusing on whole-body manipulation, a recent advance~\cite{fu2024humanplus} presents a full-stack skill transfer pipeline combining retargeting, tracking, and imitation learning, supporting learning from real-time human teleoperation data.

\subsection{Humanoid Motion Tracking}

As a core technical step for humanoids mimicking human data, humanoid motion tracking aims at controlling humanoid agents to mimic animations and generate physically realistic motions. As technical foundations, several methodologies focus on tracking small-scale animation clips from specific locomotion \cite{peng2018deepmimic,fussell2021supertrack,ren2023diffmimic}, manipulation \cite{merel2020catch,chao2021learning}, or multi-agent interaction \cite{zhang2023simulation} tasks using RL \cite{peng2018deepmimic,merel2020catch,chao2021learning,fussell2021supertrack,zhang2023simulation,yao2024moconvq} or differential simulators \cite{ren2023diffmimic}. Benefiting from large diversities of motion tasks in recent datasets \cite{ionescu2013human3,mahmood2019amass,wang2024skillmimic}, an emerging research topic is to learn universal trackers \cite{phc,serifi2024vmp,wang2023physhoi,hansen2024hierarchical} that excel in tracking versatile tasks simultaneously, which encompasses two main challenges: how to generalize the trackers to novel motions \cite{chentanez2018physics,serifi2024vmp}, and how to integrate tracking skills of numerous tasks into a single model \cite{won2020scalable,wagener2022mocapact}. As solutions to the first challenge, most existing works leverage only general reward functions (e.g., humanoid joint position rewards \cite{karim2022unicon,uhc,phc,pulse,xie2023hierarchical,he2024learning,fu2024humanplus,tevet2024closd}, task-agnostic contact graph rewards \cite{wang2023physhoi,wang2024skillmimic}) suitable for diverse motions and tasks. To handle the second challenge, a line of work 
\cite{won2020scalable,wagener2022mocapact,truong2024pdp,marew2024biomechanics} designs mixture-of-experts policies that first train specialist models to track different types of tasks respectively and then distill them to a generalist, PHC \cite{phc} and PHC+ \cite{pulse} propose active curriculum learning for hardly tracked animations, and VMP \cite{serifi2024vmp} learns motion priors covering versatile tasks and incorporates them into RL training process. With a growing demand for real-time teleoperating humanoids with sparse sensors \cite{he2024learning,ze2024generalizable,TRILL,cheng2024open,jiang2024dexmimicgen}, several recent works study tracking the animations of sparse humanoid key joints using teacher-student learning \cite{he2024omnih2o,tessler2024maskedmimic,pulse} or curriculum learning \cite{dugar2024learning} paradigms.

\section{Supplementary Experiments on Motion Tracking and Imitation Learning}
\label{supp_sec:experiments_tracking_IL}

% (optional) MaskedMimic tracking method, GAIL imitation learning method

In the main paper, we benchmark two motion tracking algorithms (Simplified PHC~\cite{phc} and Improved HST~\cite{fu2024humanplus}) and two imitation learning methods (HIT~\cite{fu2024humanplus} and ACT~\cite{act}). This section presents evaluations of several additional algorithm designs with relatively unsatisfactory performances. Section~\ref{supp_sec:experiment_tracking} examines the original HST~\cite{humanplus_codebase}, Section~\ref{supp_sec:experiment_IL} evaluates two additional imitation learning methods: Behavior Cloning (BC)~\cite{ACT_codebase} and Diffusion Policy~\cite{diffusion_policy}, and Section~\ref{supp_sec:scaling_law} reveals the data scaling law on human reference data scale for the performance improvement.

\subsection{Motion Tracking}
\label{supp_sec:experiment_tracking}

The original HST shares the same proprioception representations and training strategies with the official implementation of HumanPlus~\cite{humanplus_codebase}. Table~\ref{tab:supp_experiment_tracking} compares its performance with the two algorithms presented in the main paper, indicating the original HST gets significant performance drops on most tasks. The reason is that the proprioception representation of the original HST excludes $t$, $R$, $\bar{t}$, and $\bar{R}$ mentioned in Section~\ref{supp_sec:tracker_input} that reflect task-space joint differences, leading to missing capability 
of rectifying cumulative tracking errors in joints' global position and orientation. The task touching point (T) excludes navigation sensitive to cumulative errors, therefore the original HST achieves larger success rates on this task than on navigation-intensive SC, SS, LB, and LS.

\begin{table}[ht!]
\centering
\footnotesize
\addtolength{\tabcolsep}{-3pt}
{
\begin{tabular}{|c|cccccc|c|}
\hline
\multirow{2}{*}{Method} & \multicolumn{7}{c|}{Motion Tracking Performance} \\
\cline{2-8}
 & SC & SS & LB & LS & T & L & Mean \\
\hline
\textcolor{red}{Original HST~\cite{humanplus_codebase}} & \textcolor{red}{0/0} & \textcolor{red}{1/1} & \textcolor{red}{0/0} & \textcolor{red}{4/2} & \textcolor{red}{28/13} & \textcolor{red}{0/0} & \textcolor{red}{5/3} \\
Simplified PHC~\cite{phc} & 87/65 & \textbf{92}/83 & 63/49 & \textbf{80}/\textbf{61} & 8/6 & 0/0 & 55/44 \\
Improved HST~\cite{fu2024humanplus} & \textbf{92}/\textbf{70} & 90/\textbf{89} & \textbf{88}/\textbf{82} & 55/53 & \textbf{58}/\textbf{40} & \textbf{25}/\textbf{20} & \textbf{68}/\textbf{59} \\
\hline
\end{tabular}
}
\caption{Task success rates ($\%$) for different motion tracking methods. The retargeting algorithm is Optimization.}
\label{tab:supp_experiment_tracking}
\end{table}

\subsection{Imitation Learning}
\label{supp_sec:experiment_IL}

Behavior Cloning (BC) shares the same input and output representations with ACT, while its model is replaced with CNN-MLP architecture, which is also implemented in ACT~\cite{ACT_codebase}. Diffusion policy~\cite{diffusion_policy} is a popular method taking multiple history frames information as a part of input and learning a diffusion model to recover actions from noises conditioned on input information. Both of them are single-stage methods that directly output 19 DoF action vectors. Table~\ref{tab:supp_experiment_IL} compares their performances in task LB with the two imitation algorithms presented in the main paper, demonstrating their performances are limited in the vision-based mimicking pipeline. The reason for BC's performance is that the underlying model architecture is not as powerful as transformers in ACT to process the relations between various input modalities. For diffusion policy, one observation is that the policy is likely to give commands to complete tasks before the humanoid actually reaches a proper position for interaction.
% Hence one possible reason for diffusion policy's performance is that observation information as diffusion conditions are not strong enough to instruct the policy to leverage observations for planning.
A possible reason is that the denoised action vector in each diffusion step is dominated by the noisy action vector, which does not fully take advantage of visual observation conditions.

\begin{table}[ht!]
\centering
\footnotesize
\addtolength{\tabcolsep}{-3pt}
{
\begin{tabular}{|c|c|}
\hline
Method & Imitation Learning Performance \\
\hline
\textcolor{red}{Behavior Cloning~\cite{ACT_codebase}} & \textcolor{red}{27/27}  \\
\textcolor{red}{Diffusion Policy~\cite{diffusion_policy}} & \textcolor{red}{7/7}  \\
HIT~\cite{fu2024humanplus} & 36/34  \\
ACT~\cite{act} & \textbf{61}/\textbf{61}  \\
\hline
\end{tabular}
}
\caption{Task success rates ($\%$) for different imitation learning methods on task LB. The retargeting algorithm is Optimization and the motion tracking algorithm is the Improved HST.}
\label{tab:supp_experiment_IL}
\end{table}

\subsection{Data Scaling Law}
\label{supp_sec:scaling_law}

\begin{figure}[h!]
   \centering
   \includegraphics[width=1.0\linewidth]{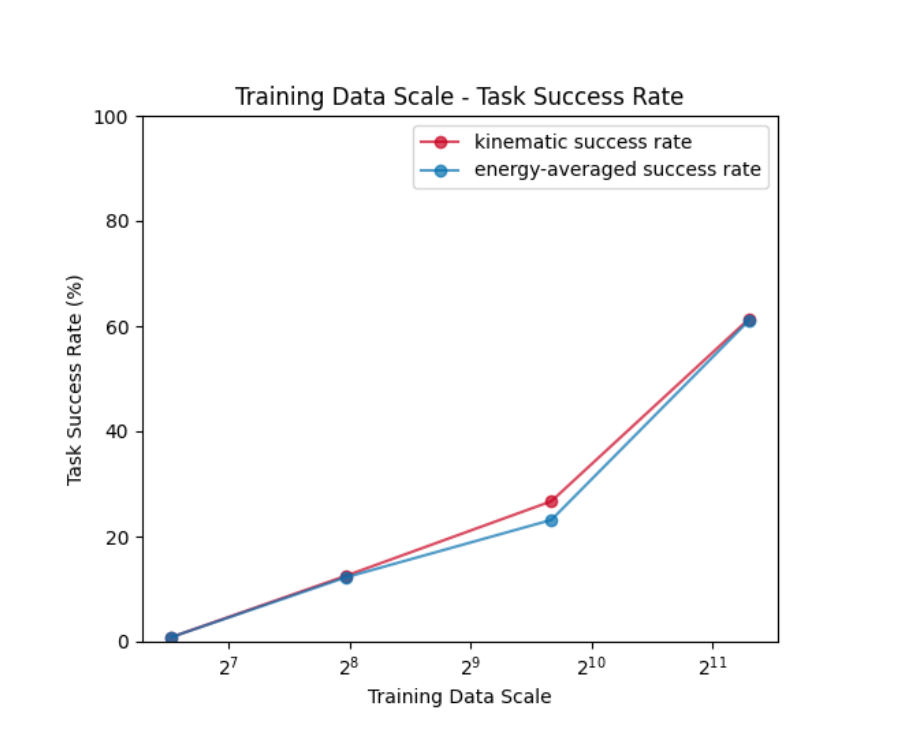}
   \vspace{-0.7cm}
   \caption{\textbf{The ascending trend of task success rates on growing training data scale.}
   }
   \vspace{-15pt}
   \label{fig:scaling_law}
\end{figure}

Section 5.6 in the main paper shows the significance of large-scale human references for imitation learning. We further compare the overall performance improvement of imitation learning policies trained on the human references of four different scales in task LB. We uncover that both kinematic and energy-averaged success rates keep an ascending trend approximately linear to the logarithm of the growing training data scale, shown in Figure~\ref{fig:scaling_law}.

\section{Details on Simulation Environment Configurations}
\label{supp_sec:simulation_enviroment}

\begin{figure}[h!]
   \centering
   \includegraphics[width=1.0\linewidth]{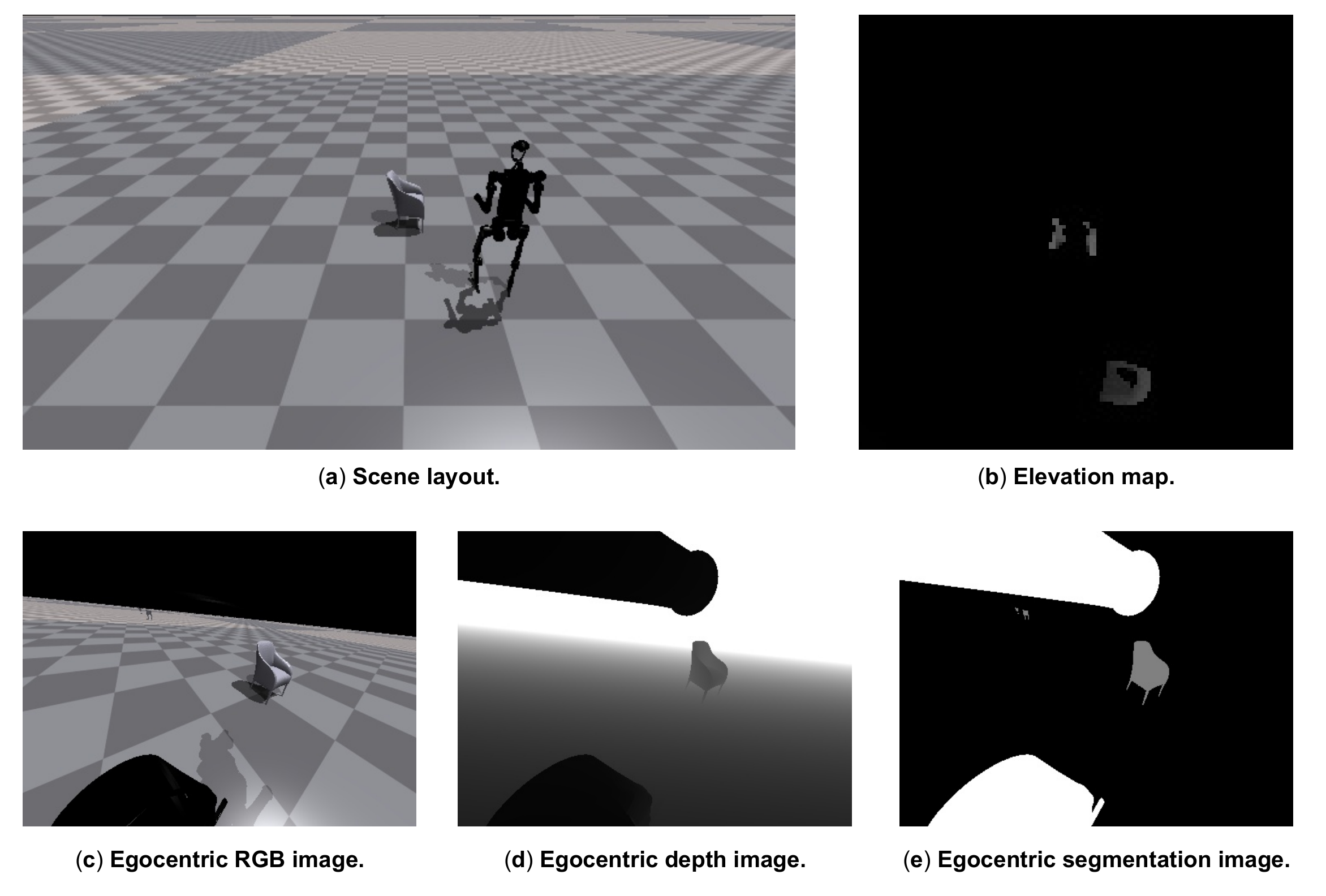}
   \vspace{-0.7cm}
   \caption{\textbf{\benchmark simulation environment configurations.} (a) exemplifies an initial scene layout for task SC. (b) shows the captured elevation map $\mathcal{E}$ in the scene. (c) (d) and (e) show the RGB image $\mathcal{C}$, depth image $\mathcal{D}$ and segmentation image $\mathcal{S}$ captured by the left front camera in the scene, respectively.
   }
   \vspace{-15pt}
   \label{fig:scene_layout_and_visual_observations}
\end{figure}

Section 3.1 in the main paper introduces our design of the simulation environment. It serves as a comprehensive platform across all six tasks to support reasonable task settings for the evaluation of the humanoid's performance on various tasks. This  section supplements details on the scene configuration (Section~\ref{supp_sec:scene_configuration})
and the setting related to visual observation
(Section~\ref{supp_sec:visual_observation}), 
as shown in Figure~\ref{fig:scene_layout_and_visual_observations}.

\subsection{Scene Configuration}
\label{supp_sec:scene_configuration}
\textbf{Initial scene layout:} 
In each simulation environment, one object with one of its corresponding motion sequences is randomly sampled from the test set mentioned in Section 5 in the main paper. The object is posed canonically at the origin of the scene and oriented towards the positive $X$-axis. The humanoid's initial pose is set to the pose in the first frame of the sampled motion sequence. For tasks SC, SS, LB, and LS that demand locating and navigating, to avoid the humanoid being too far away from the object to detect it visually, we further normalize the humanoid's initial positions beyond the range of visual detection, clipping their distances to the origin within $2\text{m}$.

\textbf{Time limit:} 
For tasks T and L, the time limit for the robot to complete the task is set to $10\text{s}$. For tasks SC, SS, LB, and LS, the time limit is set to $20\text{s}$, considering additional time cost on necessary relatively long-distance movement. 

\subsection{Egocentric Camera Setup and Visual Observation}
\label{supp_sec:visual_observation}

For each camera in all tasks, the resolution is set to $360\times 480$ with a 120-degree field of view. The methods of obtaining visual observations of the two modalities are described below.

\textbf{Camera setup for tasks SC, SS, LB, and LS:} These tasks involve complex spatial relations between the humanoid's whole body and objects. Hence in these tasks, four cameras are uniformly attached around the humanoid's pelvis and respectively set to face right front, left front, left back, and right back, all with a $35^\circ$ downward pitch. This camera configuration enables the humanoid to capture the whole surrounding environment in a 360-degree view, providing necessary self-positioning information for further planning.

\textbf{Camera setup for tasks T and L:} These tasks require precise hand motions, while their humanoid-object spatial relations are relatively simple since the humanoid initially stands in front of the object. To adapt to such requirements, we reduce the number of cameras to one to capture objects at relatively fixed positions in front of the humanoid. We locate this camera $0.55\text{m}$ above the pelvis and orient it to face forward with $55^\circ$ downward pitch, aggregating the information of arms and hands into visual inputs.

\textbf{Multi-view RGB, depth and segmentation images:} In each frame, cameras return RGB images $\mathcal{C}$, depth images $\mathcal{D}$ and segmentation images $\mathcal{S}$, with depth values clipped to $50\text{m}$. For task touching points (T), two task target hand points are projected onto RGB images, and the projected image pixels are colored red.

\textbf{Elevation map:} For visual observation $\mathcal{E}$, elevation maps~\cite{miki2022elevation} are computed from raw depth images by two steps. Depth images $\mathcal{D}$ are first filtered out depth values larger than $6\text{m}$ and remaining values are converted into point clouds $\mathcal{P}_{\text{root}}$ in the local coordinate of humanoid's root joint. Then the points in $\mathcal{P}_{\text{root}}$ are projected into uniform 2D square grids divided on the horizontal plane surrounding the humanoid. The resolution of 2D grids is $128\times128$ and each grid corresponds to a pixel in the final elevation map. For each grid $g$, the corresponding pixel value in the elevation map is the maximum height of points in $\mathcal{P}_{\text{root}}$ that are projected into $g$:

\begin{equation}
\mathcal{E}(g) = \max\left\{p_z|p\in\mathcal{P}_{\text{root}}, \text{proj}(p)\in g\right\}
\end{equation}
The final elevation map $\mathcal{E}$ is a grayscale map of resolution $128\times 128$ representing the surrounding scene height information. The humanoid is located at the center and facing downward in the elevation map. For tasks except T, the grid size is $4\text{cm}\times 4\text{cm}$. For task T, the grid size is designed as $1\text{cm}\times 1\text{cm}$ for more accurate target instruction. Besides, an additional target point elevation map $\mathcal{E}_{\text{target}}$ is provided for goal instruction, as a black image only highlighting the heights and relative positions of two target points.

\section{Details on \benchmark Task Evaluation Metrics}
\label{supp_sec:task_evaluation_metrics} 

Section 3.2 in the main paper mentions that we design the task-specific kinematic evaluation metric $K$ and the task-agnostic physical metric $P$ to examine whether a humanoid motion fulfills the task. A motion is treated as successful only if it achieves $K$ and $P$ simultaneously. This section presents details on the kinematic metrics for each task (Section~\ref{supp_sec:kinematic_metric}) and hyperparameters for the physical evaluation metric (Section~\ref{supp_physical_metric}), respectively.

\subsection{Kinematic Evaluation Metric}
\label{supp_sec:kinematic_metric}

\textbf{(1) Sitting on a Chair (SC):} $K$ evaluates whether the pelvis of the humanoid keeps above the seat surface for 0.3 seconds. In each frame, the pelvis is above the seat surface only if the root position of the humanoid lies in the chair's interior in the bird-eye view, and its height lies in $[H,H+0.27\text{m}]$, where $H$ is the height of the seat.

\textbf{(2) Sitting on a Sofa (SS):} $K$ is the same as that for SC.

\textbf{(3) Lying on a Bed (LB):} $K$ evaluates whether the pelvis and ankles of the humanoid keep on the bed for 0.3 seconds. In each frame, a humanoid joint is on the bed only if its position lies in the bed's interior in the bird-eye view, and its height lies in $[H,H+0.4\text{m}]$, where $H$ is the height of the bed's seat.

\textbf{(4) Lying on a Sofa (LS):} $K$ evaluates whether the pelvis keeps on the sofa meanwhile the ankles keep above half of the sofa's height for 0.3 seconds. In each frame, the pelvis is on the sofa only if the humanoid's root position lies in the sofa's interior in the bird-eye view, and its height lies in $[H,H+0.4\text{m}]$, where $H$ is the height of the sofa's seat.

\textbf{(5) Touching Points near an Object (T):} $K$ evaluates whether the two wrists keep within 0.1m of their target positions in 1 second.

\textbf{(6) Lifting a Box (L):} $K$ evaluates whether the box is lifted 0.2m high meanwhile the two wrists are within 0.1m of the box surface in the last frame.

\subsection{Hyperparameters on Physical Evaluation Metric}
\label{supp_physical_metric}

The physical metric $P = [\max(\Vert \tau_i v_i\Vert^2) < \lambda_P]$ evaluates whether the humanoid can keep moving with low energy cost. Due to the unknown expected value of $\lambda_P$ for real-world deployment, we empirically set $\lambda_P$ as 1e6, 2e6, 4e6, and 8e6, and report the average success rates on evaluating under these four values.

\section{Details on Motion Retargeting Algorithms}
\label{supp_sec:retargeting}

% define the mappings between each human skeleton and the humanoid, the loss function for optimization, and the bone alignment for the optimization with shape alignment.

Motion retargeting is the first step of the skill-learning paradigm, which aims to transfer motion animations from human skeletons to humanoid ones. We benchmark three types of motion retargeting algorithms: copy-rotation, optimization, and optimization with skeleton shape alignment. This section presents details on human-to-humanoid joint mapping (Section~\ref{supp_sec:retargeting_mapping}), optimization (Section~\ref{supp_sec:retargeting_optimization}), and skeleton shape alignment (Section~\ref{supp_sec:retargeting_shape_alignment}).

\subsection{Human-to-Humanoid Joint Mapping}
\label{supp_sec:retargeting_mapping}

The human-to-humanoid joint mapping $M$ builds joint-wise correspondences between the two skeletons. We provide the mapping for the three human models from UniHSI~\cite{unihsi}, ROAM~\cite{roam}, and CORE4D~\cite{core4d} in Table~\ref{tab:retargeting_mapping_unihsi},~\ref{tab:retargeting_mapping_roam}, and~\ref{tab:retargeting_mapping_core4d}, respectively. Given the joint mapping, the copy-rotation algorithm directly copies joint rotations from human to humanoid.

\begin{table}[h!]
\centering
\footnotesize
\addtolength{\tabcolsep}{-3pt}
{
\begin{tabular}{|c|c|c|c|}
\hline
\multicolumn{2}{|c|}{UniHSI~\cite{unihsi} Human Skeleton} & \multicolumn{2}{c|}{H1 Humanoid Skeleton} \\
\hline
Joint Index & Semantics & Joint Index & Semantics \\
\hline
0 & pelvis & 0 & pelvis \\
0 & pelvis & 11 & torso \\
\hline
12 & left hip & 1 & left hip yaw \\
12 & left hip & 2 & left hip row \\
13 & left knee & 4 & left knee \\
14 & left ankle & 5 & left ankle \\
\hline
9 & right hip & 6 & right hip yaw \\
9 & right hip & 7 & right hip row \\
10 & right knee & 9 & right knee \\
11 & right ankle & 10 & right ankle \\
\hline
6 & left shoulder & 13 & left shoulder row \\
7 & left elbow & 15 & left elbow \\
\hline
3 & right shoulder & 17 & right shoulder row \\
4 & right elbow & 19 & right elbow \\
\hline
\end{tabular}
}
\caption{Joint mapping between UniHSI~\cite{unihsi} human model and H1.}
\label{tab:retargeting_mapping_unihsi}
\end{table}

\begin{table}[h!]
\centering
\footnotesize
\addtolength{\tabcolsep}{-3pt}
{
\begin{tabular}{|c|c|c|c|}
\hline
\multicolumn{2}{|c|}{ROAM~\cite{roam} Human Skeleton} & \multicolumn{2}{c|}{H1 Humanoid Skeleton} \\
\hline
Joint Index & Semantics & Joint Index & Semantics \\
\hline
0 & pelvis & 0 & pelvis \\
0 & pelvis & 11 & torso \\
\hline
17 & left hip & 1 & left hip yaw \\
17 & left hip & 2 & left hip row \\
18 & left knee & 4 & left knee \\
19 & left ankle & 5 & left ankle \\
\hline
22 & right hip & 6 & right hip yaw \\
22 & right hip & 7 & right hip row \\
23 & right knee & 9 & right knee \\
24 & right ankle & 10 & right ankle \\
\hline
10 & left shoulder & 12 & left shoulder pitch \\
11 & left elbow & 15 & left elbow \\
\hline
14 & right shoulder & 16 & right shoulder pitch \\
15 & right elbow & 19 & right elbow \\
\hline
\end{tabular}
}
\caption{Joint mapping between ROAM~\cite{roam} human model and H1.}
\label{tab:retargeting_mapping_roam}
\end{table}

\begin{table}[h!]
\centering
\footnotesize
\addtolength{\tabcolsep}{-3pt}
{
\begin{tabular}{|c|c|c|c|}
\hline
\multicolumn{2}{|c|}{CORE4D~\cite{core4d} Human Skeleton} & \multicolumn{2}{c|}{H1 Humanoid Skeleton} \\
\hline
Joint Index & Semantics & Joint Index & Semantics \\
\hline
0 & pelvis & 0 & pelvis \\
0 & pelvis & 11 & torso \\
\hline
1 & left hip & 1 & left hip yaw \\
1 & left hip & 2 & left hip row \\
4 & left knee & 4 & left knee \\
7 & left ankle & 5 & left ankle \\
\hline
2 & right hip & 6 & right hip yaw \\
2 & right hip & 7 & right hip row \\
5 & right knee & 9 & right knee \\
8 & right ankle & 10 & right ankle \\
\hline
16 & left shoulder & 12 & left shoulder pitch \\
18 & left elbow & 15 & left elbow \\
\hline
17 & right shoulder & 16 & right shoulder pitch \\
19 & right elbow & 19 & right elbow \\
\hline
\end{tabular}
}
\caption{Joint mapping between CORE4D~\cite{core4d} human model and H1.}
\label{tab:retargeting_mapping_core4d}
\end{table}

\subsection{Details on Optimization}
\label{supp_sec:retargeting_optimization}

In the optimization algorithm, the state is the humanoid motion configuration $M=\{R\in\mathbb{R}^{T\times3}, t\in\mathbb{R}^{T\times3}, J\in \mathbb{R}^{T\times19}\}$, where $T$ is the frame number, $R$ is root global orientations, $t$ is root global positions, and $J$ is joint angles. We find the optimal state $M^*$ that minimizes the following loss function:

\begin{equation}
\mathcal{L_{\text{optim}}} = \lambda_{\text{pos}}\mathcal{L}_{\text{pos}} + \lambda_{\text{hand}}\mathcal{L}_{\text{hand}} + \lambda_{\text{ori}}\mathcal{L}_{\text{ori}} + \lambda_{\text{acc}}\mathcal{L}_{\text{acc}},
\end{equation}

where $\mathcal{L}_{\text{pos}}$ is the mean-square error between global positions of human and humanoid corresponding joints, $\mathcal{L}_{\text{hand}}$ is the above error between human and humanoid wrists and only applies to tasks T and L, $\mathcal{L}_{\text{ori}}$ is the mean absolute geodesic distance between global orientations of human and humanoid corresponding joints, and $\mathcal{L}_{\text{acc}}$ is the mean absolute acceleration of $J$ adopted to improve motion smoothness. $\lambda$ are hyperparameters.

The optimization is conducted by an Adam optimizer with a learning rate of 0.02 and 3000 epochs.

\subsection{Details on Motion Retargeting Algorithms}
\label{supp_sec:retargeting_shape_alignment}

OmniH2O~\cite{he2024omnih2o} proposes to find optimal SMPL-X~\cite{SMPLX} shape parameters making the human skeleton best aligned with that of H1 in the T-pose. This strategy is equivalent to adjusting human bone lengths to fit those of H1 for an arbitrary human skeleton~\cite{tang2024humanmimic}. We thus combine this strategy with the previous optimization as another type of retargeting algorithm. In practice, this strategy fits human joint global orientations and root's global position and modifies bone lengths to update human joint positions in each frame. We denote this algorithm as OmniH2O~\cite{he2024omnih2o} in experiments.

\section{Motion Tracking Algorithm Designs}
\label{supp_sec:tracking}

Given retargeted humanoid-scene interaction animations, motion tracking aims to track the animations in the simulation environment and obtain physically realistic motions as skill demonstrations. We benchmark two existing RL-based trackers, simplified PHC~\cite{phc} and improved HST~\cite{fu2024humanplus}, and present their key technical designs in this section.

\subsection{Input}
\label{supp_sec:tracker_input}

The tracker's input consists of proprioceptions $S$ of the current humanoid and the animation target of the next frame. For task touching points (T), the input also includes the wrists' position target in the current humanoid root's coordinate system. For task lifting a box (L), the input also includes states (i.e., 3D position, 3D orientation, 3D linear velocity, and 3D angular velocity) of the current box and its animation target represented in the current humanoid root's coordinate system. The proprioceptions for each tracker are defined below.

\textbf{Improved HST:} $S=\{J\in\mathbb{R}^{19}, \dot{J}\in\mathbb{R}^{19}, t\in\mathbb{R}^{20\times3}, R\in\mathbb{R}^{20\times3}, v\in\mathbb{R}^{20\times3}, \omega\in\mathbb{R}^{20\times3}, \bar{J}\in\mathbb{R}^{19}, \dot{\bar{J}}\in\mathbb{R}^{19}, \bar{t}\in\mathbb{R}^{20\times3}, \bar{R}\in\mathbb{R}^{20\times3}, \bar{v}\in\mathbb{R}^{20\times3}, \bar{\omega}\in\mathbb{R}^{20\times3}, \tilde{a}\in\mathbb{R}^{19}, g\in\mathbb{R}^3 \}$, where:

\begin{itemize}
\item $J$ and $\dot{J}$ are joint angles and velocities of the current humanoid.
\item $t$, $R$, $v$, and $\omega$ are joints' positions, orientations, linear velocities, and angular velocities of the current humanoid in the current humanoid's root coordinate system.
\item $\bar{J}$ and $\dot{\bar{J}}$ are joint angles and velocities of the animation target.
\item $\bar{t}$, $\bar{R}$, $\bar{v}$, and $\bar{\omega}$ are joints' positions, orientations, linear velocities, and angular velocities of the animation target in the current humanoid's root coordinate system.
\item $\tilde{a}$ is the generated action vector in the last frame.
\item $g$ is the gravity direction in the current humanoid's root coordinate system.
\end{itemize}

\textbf{Simplified PHC:} The $S$ for this method is the combination of the $S$ for the improved HST and $\{t-\bar{t}, R-\bar{R}\}$, which induce the tracker to learn from state differences explicitly.

\subsection{Output}
\label{supp_sec:tracker_output}

In each frame, the tracker's output is a 19 DoF action vector representing joint angles. Joint torques are then computed via predefined PD gains and fed into the simulation environment to control the humanoid physically.

\subsection{Reward Designs}

We design task-specific reward functions due to significantly different focuses on different tasks. We provide reward designs for the improved HST below, while those for the simplified PHC are mostly similar to them but exclude regularizers.

\textbf{For tasks SC, SS, LB, and LS:} The overall reward function is $r_{\text{overall}}=r_{\text{human}}+r_{\text{reg}}$, where:

\begin{itemize}
\item $r_{\text{human}}=r_{\text{pos}}+r_{\text{ori}}+r_{\text{root}}$ encourages tracker motions to approach animation targets, where $r_{\text{pos}}=\exp(-5\sum\limits_{i=1}^{20} \Vert t_i-\bar{t}_i \Vert_2^2)$ measures joint position error, $r_{\text{ori}}=\sum\limits_{i=1}^{20}\exp(-\text{Geo}(R_i, \bar{R}_i))$ measures joint orientation error (``Geo" denotes the geodesic distance between the two rotation matrices), and $r_{\text{root}}=5\exp(-10\Vert t_{\text{root}} - \bar{t}_{\text{root}} \Vert_1)-\lambda_{\text{height}} (\text{height}_{\text{root}} - \overline{\text{height}}_{\text{root}})^2$ emphasizes the root accuracy (``height" denotes the height of a joint in the world coordinate system, and $\lambda_{\text{height}}$ is set to 100 for the two sitting tasks and 10 for others).
\item $r_{\text{reg}}=r_{\text{action}}+r_{\text{vel}}+r_{\text{acc}}+r_{\text{energy}}$ is the regularizers penalizing motions with high speeds or energies, where $r_{\text{action}}=-1e-3 \Vert a - \tilde{a} \Vert_2^2$ penalizes action changes, $r_{\text{vel}}=-2e-3\Vert \dot{J} - \dot{\bar{J}} \Vert_2^2$ penalizes large joint velocities, $r_{\text{acc}}=-5e-7\Vert \ddot{J}-\ddot{\bar{J}} \Vert^2$ penalizes large joint accelerations, and $r_{\text{energy}}=1e-6\Vert \tau\dot{J} \Vert_2^2$ penalizes large joint torques and velocities.
\end{itemize}

\textbf{For task T:} To further enhance wrist accuracy, the overall reward function is $r_\text{overall}=r_{\text{human}}\times r_{\text{wrist}} + r_{\text{reg}}$. $r_{\text{human}}$ and $r_{\text{reg}}$ are similar to those for the previous four tasks, while $r_{\text{wrist}}=\exp(-10\Vert t_{\text{left wrist}} - \bar{t}_{\text{left wrist}} \Vert_1) + \exp(-10\Vert t_{\text{right wrist}} - \bar{t}_{\text{right wrist}} \Vert_1)$ highlights wrist accuracy.

\textbf{For task L:} To better induce object moving, the overall reward function is $r_\text{overall}=r_{\text{human}} \times r_{\text{w}\rightarrow\text{o}} \times r_{\text{object}} + r_{\text{reg}}$. $r_{\text{human}}$ and $r_{\text{reg}}$ are similar to those for the previous five tasks. $r_{\text{w}\rightarrow\text{o}}=\exp(-10\Vert (t_{\text{left wrist}} - t_{\text{object}}) - (\bar{t}_{\text{left wrist}} - \bar{t}_{\text{object}}) \Vert_1) + \exp(-10\Vert (t_{\text{right wrist}} - t_{\text{object}}) - (\bar{t}_{\text{right wrist}} - \bar{t}_{\text{object}}) \Vert_1)$ encourages wrist positions relative to the object to get close to those of the animation target, and $r_{\text{object}}=\exp(-10\Vert t_{\text{object}} - \bar{t}_{\text{object}} \Vert_1)$ encourages moving the object to approach its animation target. We observe that $r_{\text{human}}$, $r_{\text{w}\rightarrow\text{o}}$, and $r_{\text{object}}$ are all indispensable for completing the task successfully.

\subsection{Training Strategies}

The two trackers both include the early-termination strategy~\cite{peng2018deepmimic} that terminates the rollout when the current humanoid root is 0.5m far from that of the animation target or its height is lower than a task-specific acceptance threshold. The basic RL algorithm is ActorCritic with PPO~\cite{ppo}.

\section{Imitation Learning Algorithm Designs}
\label{supp_sec:IL}

The physically realistic motions generated by trackers can be utilized as large-scale skill demonstrations for imitation learning. The imitation learning policy aims to imitate demonstration motions during training, and output control commands to complete tasks based on proprioceptions and visual observations without any instructions from animation motions during evaluation. We benchmark two imitation learning algorithms, ACT~\cite{act} and HIT~\cite{fu2024humanplus} as representatives of the single-stage and two-stage algorithms, respectively. And we present their design details in this section.

\subsection{Input}
Compared to the tracker's input, the input of the imitation learning policy consists of proprioceptions and egocentric visual observations of the current humanoid, excluding any animation target information or direct object state information.

The definition of proprioception is the subset of $S$ in the input of the corresponding tracker (Section~\ref{supp_sec:tracker_input}) that generates demonstration data. Specifically, we partition tracker's proprioceptions $S$ into two parts $S_{\text{current}}$ and $S_{\text{target}}$, and only feed $S_{\text{current}}$ into imitation learning policies:
\begin{itemize}
\item $S_{\text{current}}=\{J, \dot{J}, t, R, v, \omega,
\tilde{a},
g \}$, which can be directly obtained in the simulation environment without animation targets.
\item $S_{\text{target}}=\{\bar{J}, \dot{\bar{J}}, \bar{t}, \bar{R}, \bar{v}, \bar{\omega}\}$, which relies on animation targets.
\end{itemize}

Egocentric visual observations are $\mathcal{C}+\mathcal{D}$ as multi-view RGBD images or $\mathcal{E}$ as elevation maps. For task touching points (T) using elevation map modality, additional target point elevation maps $\mathcal{E}_{\text{target}}$ are provided as goal instruction (Section~\ref{supp_sec:visual_observation}). Each visual observation is processed by image backbones to be further concatenated with proprioception inputs.

\subsection{Output}
The two imitation learning policies output commands for two different stages in humanoid control, which is one of the main distinctions between them. The outputs of each policy are defined below.

\textbf{ACT:} 
ACT is a single-stage algorithm and its imitation target is exactly the tracker's output. Namely, ACT aims to output the same 19 DoF action vector as the tracker outputs. 

\textbf{HIT:} 
HIT is a two-stage algorithm leveraging the corresponding tracker in inference. In the first stage of the algorithm, HIT predicts a high-level plan for target humanoid poses $\hat{S}_{\text{target}}$ in the next frame, wrists' position target if the task is touching points (T) and box states if the task is lifting a box (L). HIT's outputs complement all required input information for its corresponding tracker and are concatenated to $S_{\text{current}}$ as a tracker's complete input $\hat{S}$. In the second stage, the low-level tracker takes $\hat{S}$ and outputs the final 19 DoF action vector to physically execute the predicted plan. 

Since $S_{\text{target}}$ is in high dimension and can be recovered by its subset, instead of directly imitating it, we design HIT's imitation target of $S_{\text{target}}$ to be the subset 
$\{\bar{J}\in\mathbb{R}^{19},
\dot{\bar{J}}\in\mathbb{R}^{19}, 
\bar{t}_{\text{root}}\in\mathbb{R}^{3}, 
\bar{R}_{\text{root}}\in\mathbb{R}^{6}
\} 
$,
where $\bar{t}_{\text{root}}$, $\bar{R}_{\text{root}}$ are the root joint's position and orientation of the future humanoid in the current humanoid's root coordinate system. To recover the root velocity information, HIT predicts for the next two frames instead of the next single frame. Then all information in $S_{\text{target}}$ can be recovered by taking the temporal difference between two frames and computing forward kinematics.

\subsection{Training Strategies}
The training of the two imitation learning policies includes predictions for a future chunk instead of a single future frame to supervise the imitated control for a longer horizon. During training, HIT also supervises the difference in image embeddings between the prediction and the ground truth in a future chunk. During inference, these auxiliary supervision channels are ignored and only the predicted action in the first following frame is utilized for more precise control.

\section{Skill Learning Visualizations}
\label{supp_sec:skill_learning_visualizations}

\begin{figure*}[h!]
   \centering
   \includegraphics[width=1.0\textwidth]{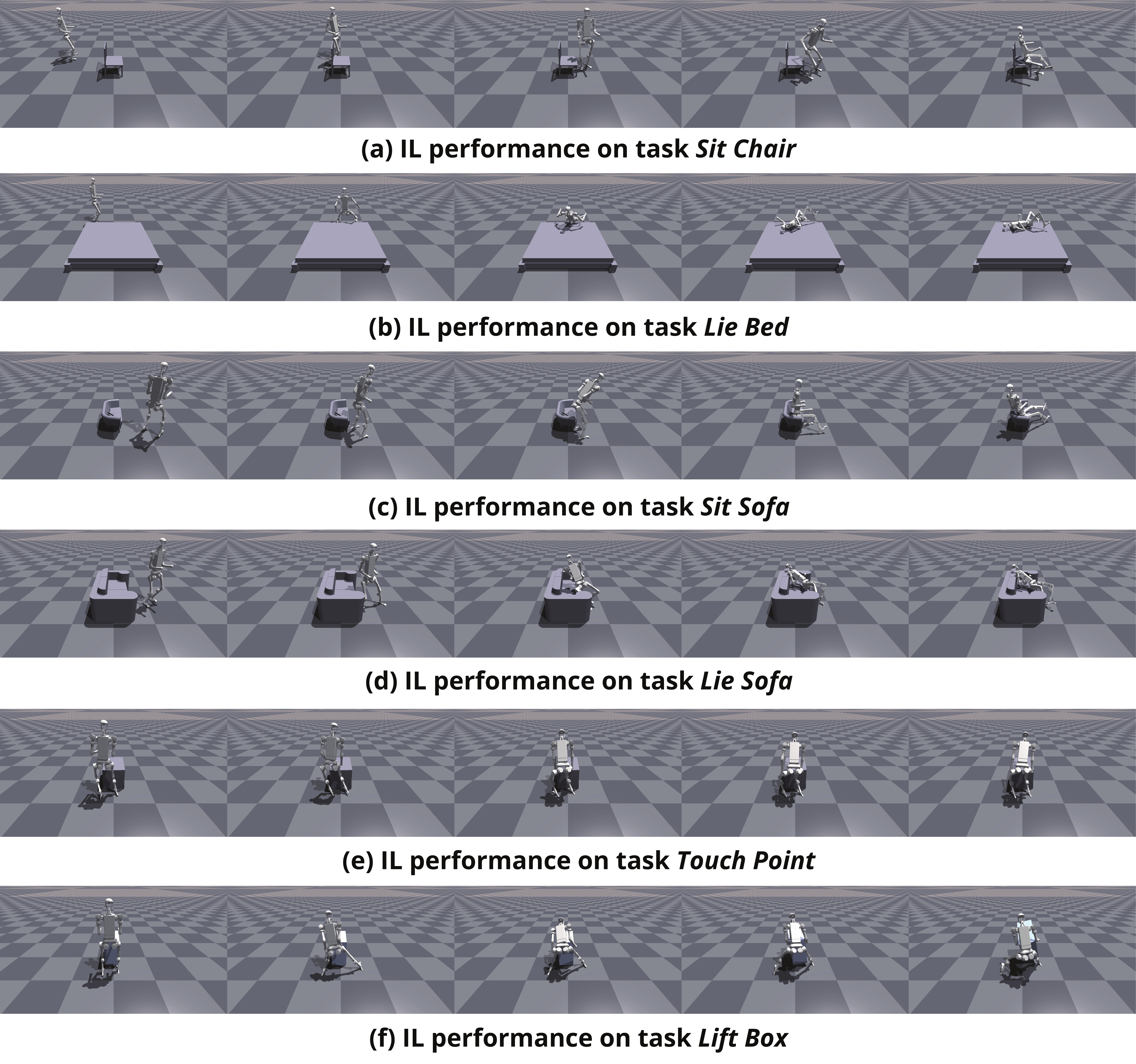}
   \vspace{-0.2cm}
   \caption{\textbf{Examples of imitation learning performance on various tasks.} The retargeting, tracking, and imitation learning algorithms are Optimization, Improved HST, and ACT, respectively.}
   \vspace{-0.4cm}
   \label{fig:IL_visualization}
\end{figure*}

Figure~\ref{fig:IL_visualization} exemplifies learned humanoid-scene interaction skills for each task. More visualizations are provided in our video.

\end{document}